\documentclass{ieeeaccess}
\usepackage[T1]{fontenc}
\usepackage{cite}
\usepackage{amsmath,amssymb,amsfonts}
\usepackage{graphicx}
\usepackage{textcomp}
\usepackage{algorithm}
\usepackage{algpseudocode}
\usepackage{booktabs}
\usepackage{subcaption}

\def\BibTeX{{\rm B\kern-.05em{\sc i\kern-.025em b}\kern-.08em
    T\kern-.1667em\lower.7ex\hbox{E}\kern-.125emX}}
\begin{document}
\history{Date of publication xxxx 00, 0000, date of current version xxxx 00, 0000.}
\doi{10.1109/ACCESS.2017.DOI}

% \captionsetup[subfigure]{labelformat=simple}
% \renewcommand{\thesubfigure}{(\alph{subfigure})}

\title{Subgoal-based Reward Shaping to Improve Efficiency in Reinforcement Learning}
% \author{}
\author{
	\uppercase{
			Takato Okudo\authorrefmark{1,2} and
			Seiji Yamada\authorrefmark{2,1}, \IEEEmembership{Member, IEEE}
	}
}
\address[1]{Department of Informatics, The Graduate University for Advanced Studies, SOKENDAI, Tokyo, 101-8430 Japan}
\address[2]{National Institute of Informatics, Tokyo 101-8430 Japan}

\markboth
{Author \headeretal: Preparation of Papers for IEEE TRANSACTIONS and JOURNALS}
{Author \headeretal: Preparation of Papers for IEEE TRANSACTIONS and JOURNALS}

\corresp{Corresponding author: Takato Okudo (e-mail: okudo@nii.ac.jp).}

\begin{abstract}
    Reinforcement learning, which acquires a policy maximizing long-term rewards, has been actively studied. Unfortunately, this learning type is too slow and difficult to use in practical situations because the state-action space becomes huge in real environments. Many studies have incorporated human knowledge into reinforcement Learning. Though human knowledge on trajectories is often used, a human could be asked to control an AI agent, which can be difficult. Knowledge on subgoals may lessen this requirement because humans need only to consider a few representative states on an optimal trajectory in their minds. The essential factor for learning efficiency is rewards. Potential-based reward shaping is a basic method for enriching rewards. However, it is often difficult to incorporate subgoals for accelerating learning over potential-based reward shaping. This is because the appropriate potentials are not intuitive for humans. We extend potential-based reward shaping and propose a subgoal-based reward shaping. The method makes it easier for human trainers to share their knowledge of subgoals. To evaluate our method, we obtained a subgoal series from participants and conducted experiments in three domains, four-rooms(discrete states and discrete actions), pinball(continuous and discrete), and picking(both continuous). We compared our method with a baseline reinforcement learning algorithm and other subgoal-based methods, including random subgoal and naive subgoal-based reward shaping. As a result, we found out that our reward shaping outperformed all other methods in learning efficiency.
\end{abstract}

\begin{keywords}
    Reinforcement Learning, Subgoals as Human Knowledge, Potential-based Reward Shaping, Reward Shaping
\end{keywords}

\titlepgskip=-15pt

\maketitle

\section{Introduction}
\label{sec:introduction}
Reinforcement learning(RL) can acquire a policy maximizing long-term rewards in an environment. Designers do not need to specify how to achieve a goal; they only need to specify what a learning agent should achieve with a reward function. A reinforcement learning agent performs both exploration and exploitation to find how to achieve a goal by itself. It is common for the state-action space to be quite large in a real environment like robotics. As the space becomes larger, the number of iterations to learn the optimal policies exponentially increases, and the learning becomes too slow to obtain the optimal policies in a realistic amount of time. Since a human could have knowledge that would be helpful to such an agent in some cases, a promising approach is utilizing human knowledge~\cite{ijcai2018-817}, \cite{DBLP:conf/atal/HarutyunyanBVN15a}, \cite{8708686}. Wang and Taylor's approach transfers a policy by requesting that a human selects an action for a state during the learning phase~\cite{DBLP:conf/aaaiss/WangT16}. Griffith et al. \cite{NIPS2013_5187} used interactive human feedback as direct policy labels.\par
The reward function is the most related to learning efficiency. Most difficult tasks in RL have a sparse reward function~\cite{Yusuf2018}.
The agent is not able to evaluate its policy due to it and to learn the optimal policy. In contrast, learning speeds up when the reward function is dense. Inverse reinforcement learning~(IRL)~\cite{Ng:2000:AIR:645529.657801}, \cite{10.1145/1015330.1015430} is the most popular method for enriching the reward function. IRL uses an optimal policy to generate a dense reward function. 
Recent studies have utilized optimal trajectories~\cite{NIPS2016_6391}, ~\cite{10.5555/1620270.1620297}. 
There is the question of the cost for the teacher in providing the optimal trajectories or policies. Humans sometimes have difficulty providing these because of the skills they may or may not have. In particular, in robotics tasks, humans are required to have robot-handling skills and knowledge on the optimal trajectory. 
We focus on the knowledge of subgoals because there are no requirements for robot-handling skills. Humans need only to think of a few subgoal states on the optimal trajectory in their minds. Moreover, trajectories that fail can be utilized. Potential-based reward shaping is able to add external rewards while keeping the optimal policy of the environment~\cite{Ng+HR:1999}.
It is calculated as the difference between the real-number functions~(potential function) of the previous and current state. Generally, a naive idea of the potential function for utilizing the knowledge of subgoals is high potential only for subgoal states. However, the method cannot work well as shown in Section~\ref{sec:eval}, so it is not easy to incorporate the knowledge of subgoals over potential-based reward shaping. \par
In this paper, we propose a subgoal-based reward shaping, that accelerates learning with the knowledge of subgoals with the original optimal policy remaining. 
The remainder of this paper is organized as the followings. Section II reviews related work on RL and similar studies to our reward shaping with subgoal knowledge. 
The preliminaries including notations and algorithms for RL and potential based reward shaping are described in Section \ref{sec:preliminaries}. Then, we explain the definition of subgoals, a subgoal-based reward shaping algorithm we proposed in Section \ref{sec:srs}. The next Section \ref{sec:user_study}  gives an overview of user study in which we obtained  subgoal knowledge from participants. The setting and results of various experiments we conducted for evaluation of our proposed method and descriptions on three domains, four-rooms, pinball, pick and place tasks in Section \ref{sec:eval}. In Section \ref{sec:discussion}, discussions on subgoals given by participants, hyper parameters tuning and others are discussed. The conclusion is provided in Section \ref{sec:conclusion}.

\section{Related Work}
% 全体の流れ
% Human Knowledge (Preference learning, Inverse RL, DAGGER)
\subsection{Human Knowledge for Reinforcement Learning}
There are many studies utilizing human knowledge to RL; trajectory, policy, preference, action, feedback, and subgoal. 
% Trajectory
From policy or trajectory which human provides, IRL infers an unobserved reward function~\cite{ARORA2021103500}. It is known to be difficult to design the reward functions because of an unspecified reward~\cite{DBLP:journals/corr/AmodeiOSCSM16} and a reward hacking~\cite{10.5555/1671238}. The unspecified reward is leaving out important aspects in reward design. The reward hacking is short of reward design for penalizing all possible misbehavior. Since the designer only provide the policy or trajectories without defining the reward function directly, the IRL overcomes these difficulties. The reward function is modeled as a linear sum of weighted features, and is acquired by a margin-based optimization~\cite{Ng:2000:AIR:645529.657801} or an entropy-based optimization~\cite{NIPS2016_6391}. Since the IRL generates a rich reward function, it is also a powerful solution when an environment has a sparse reward. The approach is same to ours. The difference is that we use only some specific states as subgoal. \par
% Preference
In RL with human preferences~\cite{Christiano2017}, a human teacher compares a pair of trajectory segments and select the better one than the other one. The method learned a policy without access to the environmental rewards and the human preference predictor. \par
% Dataset Aggregation
In dataset aggregation(DAGGER)~\cite{pmlr-v15-ross11a}, a human expert selects an action in a state which an agent selects.  \par
% Interactive RL, feedback
In the field of interactive reinforcement learning, a learning agent interacts with a human trainer as well as the environment\cite{doi:10.1080/09540091.2018.1443318}. The TAMER framework is a typical interactive framework for reinforcement learning\cite{KCAP09-knox}. The human trainer observes the agent's actions and provides binary feedback during learning. Since humans often do not have programming skills and knowledge on algorithms, the method relaxes the requirements to be a trainer. We aim for fewer trainer requirements, and we use a GUI on a web system in experiments with navigation tasks.  \par
% Human Subgoal（RGoal Arch, Subgoal induction）
RGoal architecture by Ichisugi et al.~\cite{10.1007/978-3-030-30484-3_9} deals with recursive subgoals in HRL. The method solves a MDP with an augmented state space in which each subgoal has a state space. In multitask settings, the method sped up the learning by sharing subroutines between tasks. Murata et al.~\cite{10.5555/1295303.1295360} used subgoals for accelerating learning in RL. A reward is generated at achievement of each subgoal, there is a $Q$-function for each subgoal. The policy select an action according to the weighted sum of $Q$-functions of subgoals and an original $Q$-function.
\par
\subsection{Subgoals for Reinforcement Learning}
% Random HER
We consider the hindsight experience replay(HER) algorithm~\cite{NIPS2017_7090}, which uses subgoals. HER regards a failure as a success in hindsight, and generates a reward for the failure. The algorithm interprets a failure as the process to achieve target, and it is really similar to a subgoal. HER accelerate the learning in a multi-goal environment. We focus on a single-goal environment.
\par
% Hierarchical RL
Hierarchical reinforcement learning(HRL) utilizes subgoals. The option framework is major in the field of HRL. The framework of Sutton et al.~\cite{Sutton:1999:MSF:319103.319108} was able to transfer learned policies in an option. An option consists of an initiation set, an intra-option policy, and a termination function. An option expresses a combination of a subtask and a policy for it. The termination function takes on the role of subgoal because it terminates an option and triggers the switching to another option. Recent methods have found good subgoals for a learner simultaneously with policy learning~\cite{Bacon:2017:OA:3298483.3298491}, \cite{10.5555/3305890.3306047}. The differences with our method are whether the policy is over abstract states or not and whether rewards are generated. The framework intends to recycle a learned policy, but our method focuses on improving learning efficiency. \par
\subsection{Reward Shaping}
% Reward Shapingの研究について
Reward shaping has been studied actively. Marom and Rosman~\cite{Marom2018} proposed the reward shaping based on Bayesian methods called belief reward shaping(BRS). The expected value over the estimated probability distribution of an environmental reward over a state and an action is used as shaping rewards. They proved that the policy learned by Q-learning augmented with BRS is consistent to an optimal policy over original MDP. \par
% Landmark-based Reward Shaping
The landmark-based reward shaping of Demir et al.~\cite{Demir2019} is the closest to our method. The method shapes only rewards on a landmark using a value function. Their study focused on a POMDP environment, and landmarks automatically become abstract states. We focus on a Markov decision process (MDP) environment, and acquire subgoals from human participants.  \par
% Reward shaping for HRL
Reward shaping in HRL has been studied in~\cite{Gao2015, Li2019}. Gao et al.~\cite{Gao2015} showed that potential-based reward shaping remains policy invariant to the MAX-Q algorithm. Designing potentials every level is laborious work. We use a single high-level value function as a potential, which reduces the design load. Li et al.~\cite{Li2019} incorporated an advantage function in high-level state-action space into reward shaping. The reward shaping method in~\cite{Paul2019} utilized subgoals that are automatically discovered with expert trajectories. The potentials generated every subgoal are different. \par
% Potential-based Advice
Potential-based advice is reward shaping for states and actions\cite{Wiewiora2003}. The method shapes the Q-value function directly for a state and an action, and it makes it easy for a human to give advice to an agent regarding whether an action in an arbitrary state is better or not. Subgoals show what ought to be achieved on the trajectory to a goal. We adopted the shaping of a state value function. \par
% Expressing Arbitrary Reward Functions as Potential-Based Advice
Harutyunyan et al.~\cite{Harutyunyan2015} has shown that the Q-values learned by arbitrary rewards can be used for potential-based advice. The method mainly assumes that a teacher negates the agent's selected action. The method uses failures in trial and errors. In contrast, our method uses successes. 

\section{Preliminaries}
\label{sec:preliminaries}
\subsection{Reinforcement Learning}
Reinforcement learning is a framework for acquiring a policy maximizing future rewards through interactions between an agent and an environment. This works under a Markov decision process(MDP). A MDP is represented as a tuple $(S, A, T, \gamma, R)$, where $S$ is a finite set of states, $A$ is a set of actions of $k\geq2$, $T = P(s_{t+1}| s_t, a)$ is a state transition function, $\gamma \in \left(0, 1\right]$ is the discount factor, and $R$ specifies a reward function. An agent has a policy $\pi(a|s)$ and a value function $V(s)$, $Q(s,a)$. There are two methods for learning a optimal policy, the value-based method \cite{Szepesvari:2010:ARL:1855083} and the policy gradient-based method \cite{NIPS1999_1713}. The value-based method indirectly learns a policy by estimating the optimal value function. The policy gradient-based method directly learns the parameters of a policy. \par
% We explain the value-based method, Q-learning, in this section. In a learning experiment, we used a hybrid method of value based and policy gradient, the actor-critic algorithm \cite{NIPS1999_1786}. \par
The initial state is $s_0 \in S$. The agent selects action $a \in A$ in the current state $s_t$, and the environment then returns the next state $s_{t+1}$ and reward $r$. The agent updates policy $\pi(a|s)$ on the basis of reward $r$. Note that $S$ is a state space and that $A$ is an action space. The value function is defined as follows,

\begin{eqnarray}
	\label{eq:vfunc}
	V_\pi(s_t) &=& \mathbb{E}_{\pi} \left[ \sum_{i=t}^{\infty} \gamma^{i-1} r_i |s_t=s_t \right] \\
	\label{eq:qfunc}
	Q_\pi(s_t,a) &=& \mathbb{E}_{\pi} \left[ \sum_{i=t}^{\infty} \gamma^{i-1} r_i |s_t=s_t, a=a \right]
\end{eqnarray}
The above equations (\ref{eq:vfunc}) and (\ref{eq:qfunc}) show an expected discounted cumulative reward achieved by following policy $\pi$ when an agent is in a state or a state-action. $\gamma$ is a discount rate, $V$ is a value function and $Q$ is an action value function. The optimal values are shown as follows.
\begin{eqnarray}
	V_\pi^*(s_t) &=& \max_{\pi} V_{\pi}(s_t) \nonumber \\
	&=& \max_{a'} Q^*(s_t, a') \\
	Q_\pi^*(s_t, a) &=& \max_{\pi} Q_{\pi}(s_t, a) \nonumber \\
	&=& r_t + V^*(s_{t+1}) \nonumber \\
	&=& r_t + \max_{a'} Q^*(s_{t+1}, a')
\end{eqnarray}

The agent does not know the dynamics of an environment like the state transition function and reward function. The agent needs to estimate them by sampling. Q-learning is a core reinforcement learning method for estimating the optimal action value function~\cite{Sutton1998}. The update equation is as follows.

\begin{equation}
	\label{q-learn}
	\begin{split}
		TD = r + \gamma \max_{a'} Q_\pi(s_{t+1}, a') - Q_\pi(s_t, a) \\
		Q_\pi(s_t, a) \leftarrow Q_\pi(s_t, a) + \alpha TD
	\end{split}
\end{equation}
$\alpha$ is the learning rate. The action value function converges to the optimal one under $0 < \alpha < 1$ and an infinite number of trials. 

\subsection{Potential based Reward Shaping}
Potential based reward shaping (PBRS)~\cite{Ng+HR:1999, DBLP:journals/corr/abs-1106-5267} is an effective method for keeping an original optimal policy in an environment with an additional reward function. If the potential-based shaping function $F$ is formed as 
\begin{eqnarray}
	F(s_t,s_{t+1}) = \gamma \Phi(s_{t+1}) - \Phi(s_t)
\end{eqnarray}
it is guaranteed that policies in MDP $M=(S,A,T, \gamma, R)$ are consistent with those in MDP $M'=(S,A,T,\gamma, R+F)$. $\Phi(s_t)$ is a function with a state $s_t$ as an argument.
Wiewiora showed that a reinforcement learner with initial $Q$-values based on the potential function make the same updates throughout learning as a learner receiving potential based shaping reward~\cite{DBLP:journals/corr/abs-1106-5267}. We assume two learners $L$ and $L'$. $L$ initializes the $Q$-value to $Q_0(s,a)$ and updates its policy with an environmental reward $r$ and the shaping reward $F$. The $Q$-value of $L'$ is initialized to $Q'_0(s,a) = Q_0(s,a)+\Phi(s)$, and $L'$ uses only the reward $r$. The $Q$-values are disassembled into the initial values $Q_0$ and $Q'_0$ and the updated values $\Delta Q(s,a)$ and $\Delta Q'(s,a)$ through learning, respectively. If $L$ and $L'$ learn on the same sequence of experiences, $\Delta Q(s,a)$ is always equal to $\Delta Q'(s,a)$. Moreover, value-based policies of $L$ and $L'$ have an identical probability distribution for their next action. Our proposed method uses potential-based shaping rewards to keep an optimal policy in an original MDP. 

\section{Subgoal-based Reward Shaping}
\label{sec:srs}
We propose subgoal-based reward shaping(SRS), which incorporates human knowledge of subgoals into an algorithm via reward shaping. In \cite{Ng+HR:1999}, it was mentioned that learning sped up with a learned value function used as the potential function. Since a policy learned with potential-based reward shaping is equivalent to that with $Q$-value initialization with the potential function~\cite{DBLP:journals/corr/abs-1106-5267}, using the learned value function for the potential function is equivalent to initializing the value function with the learned value function. It is impossible to prepare the learned values before learning. Therefore, we consider an approximation of the optimal values by using subgoals without a learned value function. To this end, we define the subgoal first, and we extend the potential-based reward shaping, and then we propose a subgoal-based potential in this section.

\subsection{Subgoal}
\begin{figure}
    \centering
		\begin{subfigure}[b]{\linewidth}
			\centering
			\includegraphics[width=\linewidth]{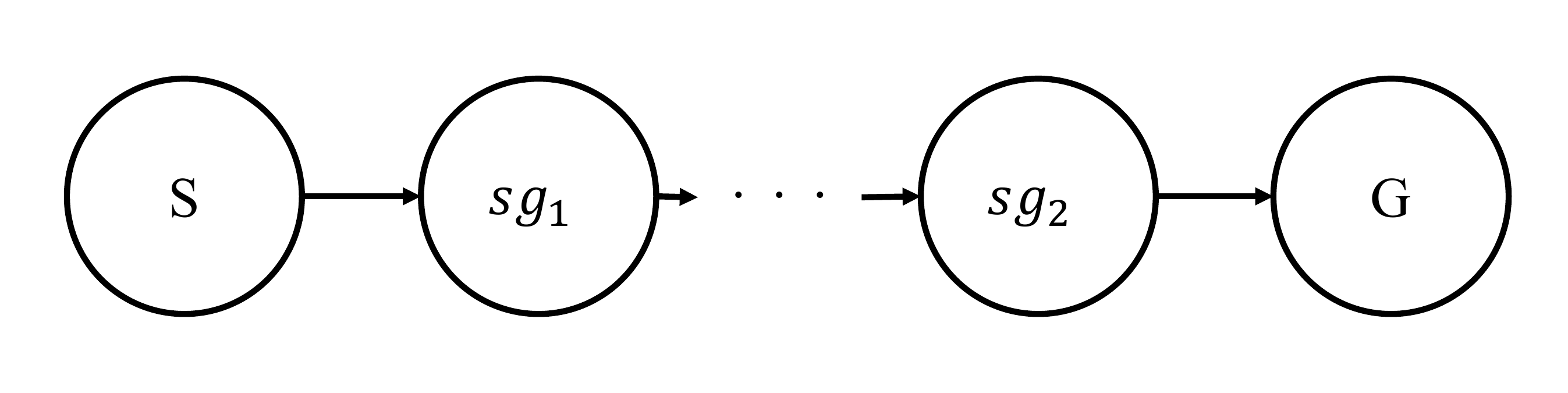}
			\caption{A totally one.}
			\label{fig:totally-ordered}
	\end{subfigure}
	\hfill
	\begin{subfigure}[b]{\linewidth}
			\centering
			\includegraphics[width=\linewidth]{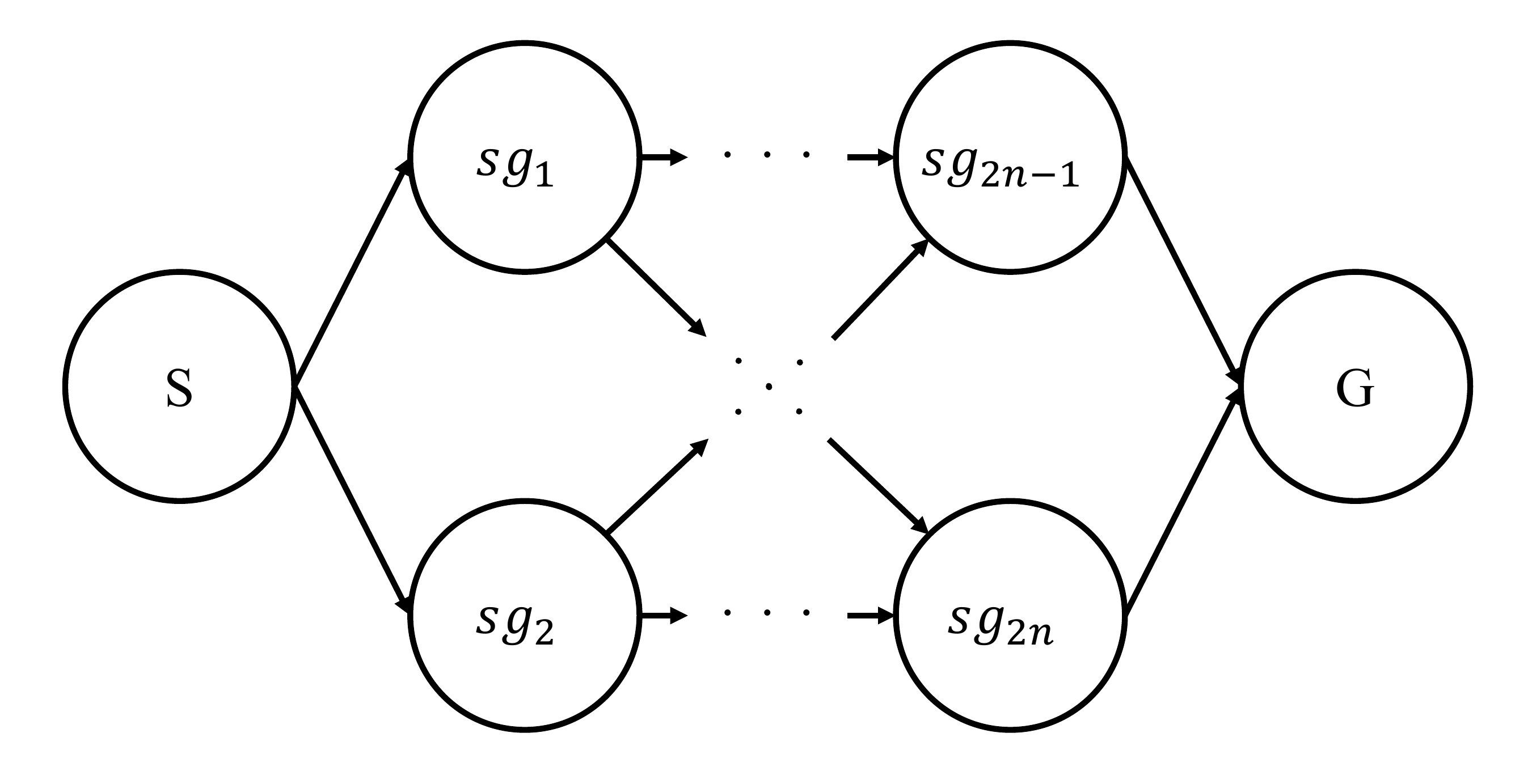}
			\caption{A partially one.}
			\label{fig:partially-ordered}
	\end{subfigure}
	\caption{Ordered subgoals.}
	\label{fig:ordered_subgs}
\end{figure}

We define a {\it subgoal} as a state $s$ if $s$ is a goal in one of the sub-tasks decomposed from a task. In the option framework, the subgoal is the goal of a sub-task, and it is expressed as a termination function~\cite{Sutton:1999:MSF:319103.319108}. Many studies on the option framework have developed automatic subgoal discovery~\cite{Bacon:2017:OA:3298483.3298491}. We aim to incorporate human subgoal knowledge into the reinforcement learning algorithm with less human effort required. The property of a subgoal might be a part of the optimal trajectories because a human should decompose a task to achieve a goal. We acquire a subgoal series and incorporate subgoals into our method in the experiments. The subgoal series is written formally as $(SG, \prec)$. $SG$ is a set of subgoals and a sub-set of $S$. 
There are two types of subgoal series, totally ordered and partially ordered as shown as Fig.~\ref{fig:ordered_subgs}.

With totally ordered subgoals, a subgoal series is deterministically determined at any subgoal. In contrast, partially ordered subgoals have several transitions to the subgoal series from a subgoal. We used only the totally ordered subgoal series in this paper, but both types of ordered subgoals can be used with our proposed reward shaping. Since an agent needs to achieve a subgoal only once, the transition between subgoals is unidirectional. 

\subsection{Potential based Reward Shaping wit State History Guarantees the Policy Invariance}
We extended potential-based reward shaping to allow the history of the states an agent visited as an input of a potential function. Informally, if the potential function obtains the history of states in the difference equation, the guarantees of policy invariance remain. Formally,

\begin{eqnarray}
\label{eq:spbrs}
F(h_t, h_{t+1}) = \gamma \Phi(h_{t+1}) - \Phi(h_t)
\end{eqnarray}

where $h_t$ is the history of the states that an agent visited until state $s_t$, expressed as an equation, $h_t = [s_0, \cdots, s_t]$. To prove the policy invariance, we first define the return in any arbitrary policy $\pi$ when visiting a state $s_k$ in a discount manner without shaping. Formally,
\begin{equation}
U_\pi(s_k) = \sum_{t=0}^{\infty} \gamma^tr_{t,\pi}
\end{equation}
Note that $t$ is a time step. The Q function can be rewritten as follows:
\begin{eqnarray}
\label{eq:extend-qfunc}
Q_{\pi}(s_{k-1},a) = \sum_{s_{k}}P(s_{k}|s_{k-1},a)U_{\pi}(s_{k})
\end{eqnarray}
Now consider the same policy but with a reward function modified by adding a  potential-based reward function of the form given in Equation (\ref{eq:spbrs}). The return of the shaped policy $U_{\pi, \Phi}$ experiencing the same sequence $s_k$ is,
\begin{eqnarray}
U_{\pi, \Phi}(h_k) &=& \sum_{t=0}^{\infty} \gamma^t \left( r_{t,\pi} + F(h_{t+k}, h_{t+k-1}) \right) \nonumber \\
&=& \sum_{t=0}^{\infty}\gamma^t \left( r_{t,\pi} + \gamma\Phi(h_{t+k}) - \Phi(h_{t+k-1}) \right) \nonumber \\
% &=& \sum_{t=0}^{\infty}\gamma^t r_{t,\pi} + \sum_{t=0}^{\infty}\gamma^{t+1}\Phi(h_{t+k}) - \sum_{t=0}^{\infty}\gamma^t\Phi(h_{t+k-1}) \nonumber \\
&=& \sum_{t=0}^{\infty}\gamma^t r_{t,\pi} - \Phi(h_{k-1}) \nonumber \\
\label{eq:return}
&=& U_{\pi}(s_k) - \Phi(h_{k-1})
\end{eqnarray}

where $h_k = [s_0,\cdots,s_k]$ is written. $Q_{\pi, \Phi}$ follows from Equation (\ref{eq:extend-qfunc}) and is expressed as follows.

\begin{eqnarray}
Q_{\pi, \Phi}(s_{k-1},a) \!\!\!&=& \!\!\!\sum_{s_{k}}P(s_k|s_{k-1},a)U_{\pi, \Phi}(h_{k})  \nonumber \\
&=&\!\!\! \sum_{s_{k}}P(s_k|s_{k-1},a) \left( U_{\pi}(s_k) - \Phi(h_{k-1}) \right) \nonumber \\
&=&\!\!\!\sum_{s_{k}}P(s_k|s_{k-1},a) U_{\pi}(s_k) \nonumber \\
&\quad&\!\!\! - \sum_{s_{k}}P(s_k|s_{k-1},a) \Phi(h_{k-1})  \nonumber \\
\label{eq:potential-qfunc}
&=&\!\!\!Q_{\pi}(s_{k-1},a) - \Phi(h_{k-1}) 
\end{eqnarray}

Given by Equation (\ref{eq:potential-qfunc}), $\Phi$ does not have action $a$ in parameters. As $\Phi$ is independent of the action taken, in any given state, the best action remains constant regardless of shaping. Therefore, we can conclude that the guarantee of policy invariance remains.

\subsection{Subgoal-based Potentials}
\begin{figure}[tb]
	\includegraphics[width=\linewidth]{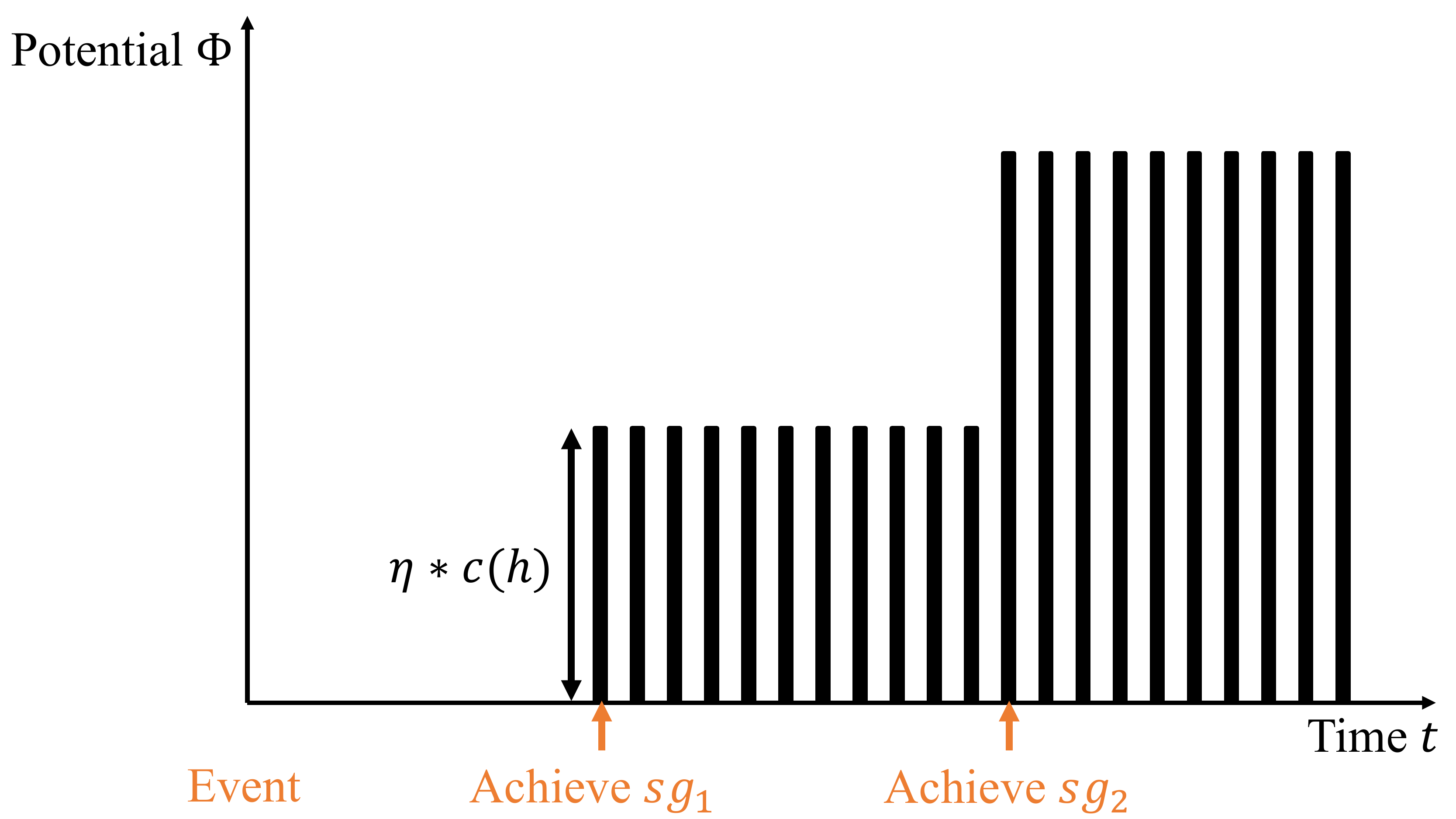}
	\caption{Potential function for subgoals.}
	\label{fig:potential-func}
\end{figure}

The subgoal-based potential is the main part of our method. We design potentials that utilize subgoals to approximate of the optimal value functions. Generally, the larger the optimal value is, the closer a state is in an environment with only goal reward. We consider subgoal achievement as approaching the goal, and we design a potential that grows every subgoal achievement. Formally, we define subgoal-based potentials as

\begin{equation}
    \label{subgoal-potential}
    \Phi(h) = \eta * c(h)
\end{equation}

where $\eta$ is a hyper parameter, and $c(h)$ is a function for calculating the number of achieved subgoals in a visited state sequence of $h$ from the last achieved subgoal. 
An output sequence of the potential function is shown in Fig.~\ref{fig:potential-func}.

Fig.~\ref{fig:potential-func} shows the sequence of an agent achieving three subgoals. When the agent achieves $sg_1$, the potential function generates $\eta * c(h)$. Upon the achievement of $sg_2$, this value becomes larger than $sg_1$ because $c(h)$ increases.

\subsection{Whole Algorithm}
The pseudocode is shown in Algorithm \ref{alg:transfer_subgoals}:
\begin{algorithm}
	\caption{Subgoal potential-based reward shaping.} 
	\label{alg:transfer_subgoals}
	\begin{algorithmic}[1]
		\Ensure $\gamma,\eta, SG$
		\State $s \leftarrow s_0, h \leftarrow [], h' \leftarrow []$
		\Repeat
		\State Choose $a$ according to $\pi$
		\State Take $a$ in $s$, observe $s'$, $r$
		\State $h' \leftarrow h$
		\State Append $s' $ into $h'$ 
		\State $\Phi(h') = \eta * c(h')$
		\State $F(h,h') = \gamma\Phi(h') - \Phi(h)$
		\State Update value function and policy with $F(h,h')$
		\State $s \leftarrow s', h \leftarrow h'$
		\Until{termination.}
	\end{algorithmic}
\end{algorithm}

Though we consider totally ordered subgoals in the experiments, extending the algorithm with a sequence of partially ordered subgoals is easy. We modify only the function~$c(h)$ for partially ordered subgoals. After an agent takes an action $a$ and receives a state $s'$ from an environment, add the state $s'$ to the history of states~$h'$. The calculation of $\Phi$ involves $h$, and $h'$. The function $\Phi(h)$ is used for computing $F(h, h')$.

\section{User Study: Human Subgoal Acquisition}
\label{sec:user_study}
In this section, we explain a user study done to acquire human knowledge of subgoals. We used navigation tasks in two domains, four-rooms and pinball, that express the state space spatially. Additionally, we conducted a pick and place task, which is a basic robotics task and one for which humans can have difficulty controlling the robot.

\subsection{Navigation Task}
\begin{figure}[tb]
    \centering
    \includegraphics[width=\linewidth]{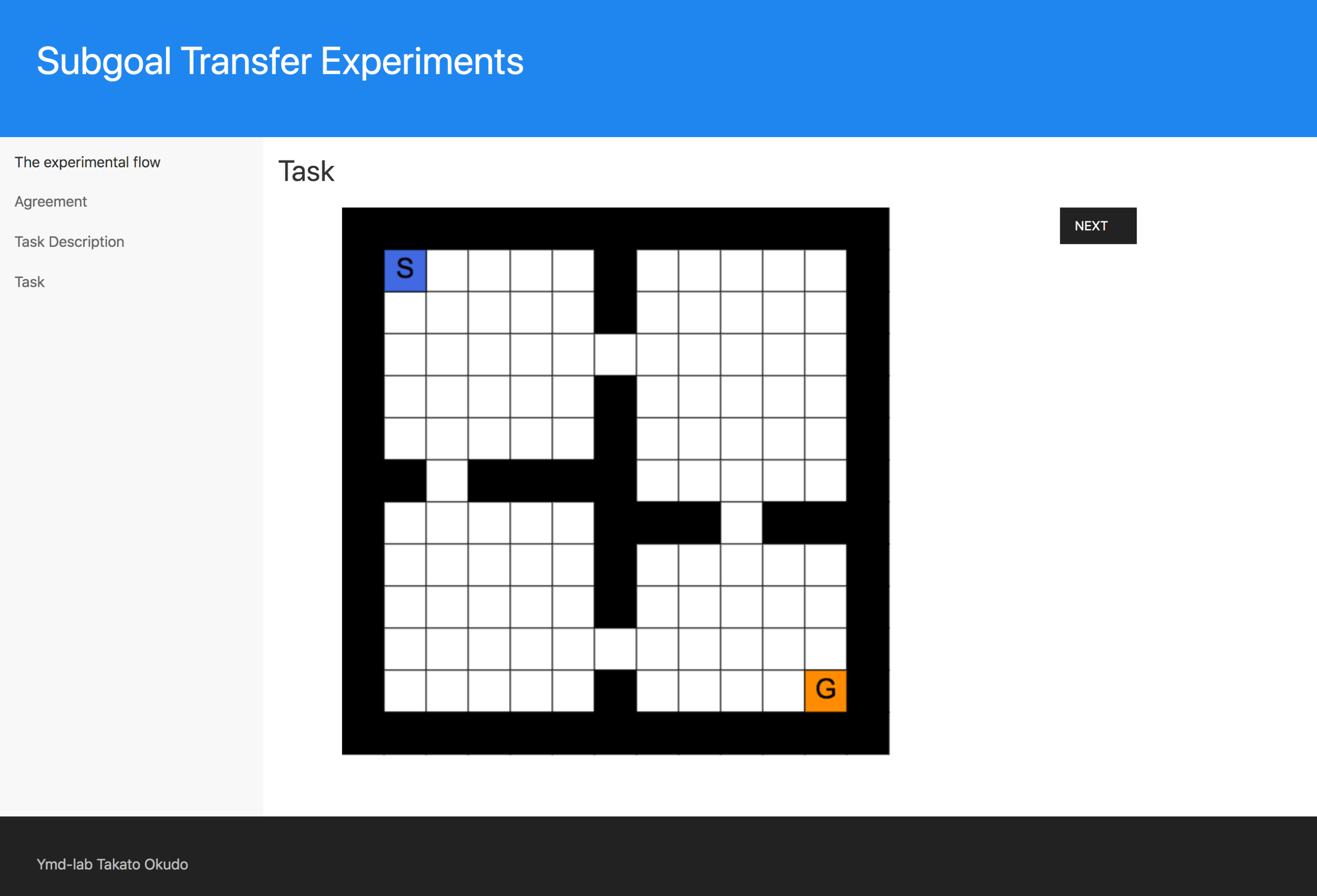}
    \caption{UI for acquiring subgoals.}
    \label{fig:ui}
\end{figure}

An online user study done to acquire human subgoal knowledge using a web-based GUI in Fig.~\ref{fig:ui} was conducted.

% TODO UIの説明

We recruited 10 participants who consisted of half graduate students in the department of computer science and half others(6 males and 4 females, ages 23 to 60, average of 36.4). We confirmed they did not have expertise on subgoals in the two domains. Participants were given the same instructions as follows for the two domains, and they were then asked to designate their two subgoals both for the four-rooms and pinball domains in this fixed order. The number of subgoals was the same as the hallways in the optimal trajectory for the four-rooms domain.
The instructions explained to the participants what the subgoals were and how to set them through the GUI shown in Fig.~\ref{fig:ui}. Also, specific explanations of the two task domains were given to the participants. In this experiment, we acquired just two subgoals for learning since they are intuitively considered easy to give on the basis of the structure of the problems. We considered the two subgoals to be totally ordered ones. \par
% Fig.~\ref{fig:subgoal-dist} shows the subgoal distribution acquired from the participants.

% \begin{figure}[tb]
%     \begin{subfigure}[b]{\linewidth}
%         \centering
%         \includegraphics[width=0.5\linewidth]{misc/4rooms-2sg-human.pdf}
%         \caption{Four-rooms domain}
%         \label{fig:4rooms_subg_dist}
%     \end{subfigure}
%     \hfill
%     \begin{subfigure}[b]{\linewidth}
%         \centering
%         \includegraphics[width=0.5\linewidth]{misc/pinball-2sg-human.pdf}
%         \caption{Pinball domain}
%         \label{fig:pinball_subg_dist}
%     \end{subfigure}
%     \caption{Participants' subgoal distribution}
% 	\label{fig:subgoal-dist}
% \end{figure}

% The numbers in the cells stand for the frequencies of the subgoals in Fig.~\ref{fig:4rooms_subg_dist}; higher values are in yellow and lower values are in orange. The number in a cell was the frequency at which the participants selected that cell as a subgoal.  These subgoals included totally and partially ordered subgoals. As shown in Fig.~\ref{fig:4rooms_subg_dist}, participants tended to set the subgoals in the hallways. 
% In Fig.~\ref{fig:pinball_subg_dist}, the color of the start point, the goal, and subgoals are red, blue, and green, respectively.

\subsection{Pick and Place Task}
A user study for the pick and place task was also done online. Since it was difficult  to acquire human subgoal knowledge with a GUI, we used a descriptive answer-type form. We assumed that humans use subgoals when they teach behavior in a verbal fashion. They state not how to move but what to achieve in the middle of behavior. The results of this paper minorly support this assumption. We recruited five participants who were amateurs in the field of computer science(3 males and 2 females, ages 23 to 61, average of 38.4). The participants read the instructions and then typed the answer in a web form. The instructions consisted of a description of the pick and place task, a movie of manipulator failures, a glossary, and a question on how a human can teach successful behavior. The question included the sentence ``Please teach behavior like you would teach your child.” This is because some participants answered that they did not know how to teach the robot in a preliminary experiment. We imposed no limit on the number of subgoals.\par

\section{Experiments for Evaluation}
\label{sec:eval}
\begin{figure}[tb]
	\centering
	\begin{subfigure}[b]{0.49\linewidth}
		\centering
		\includegraphics[width=\linewidth]{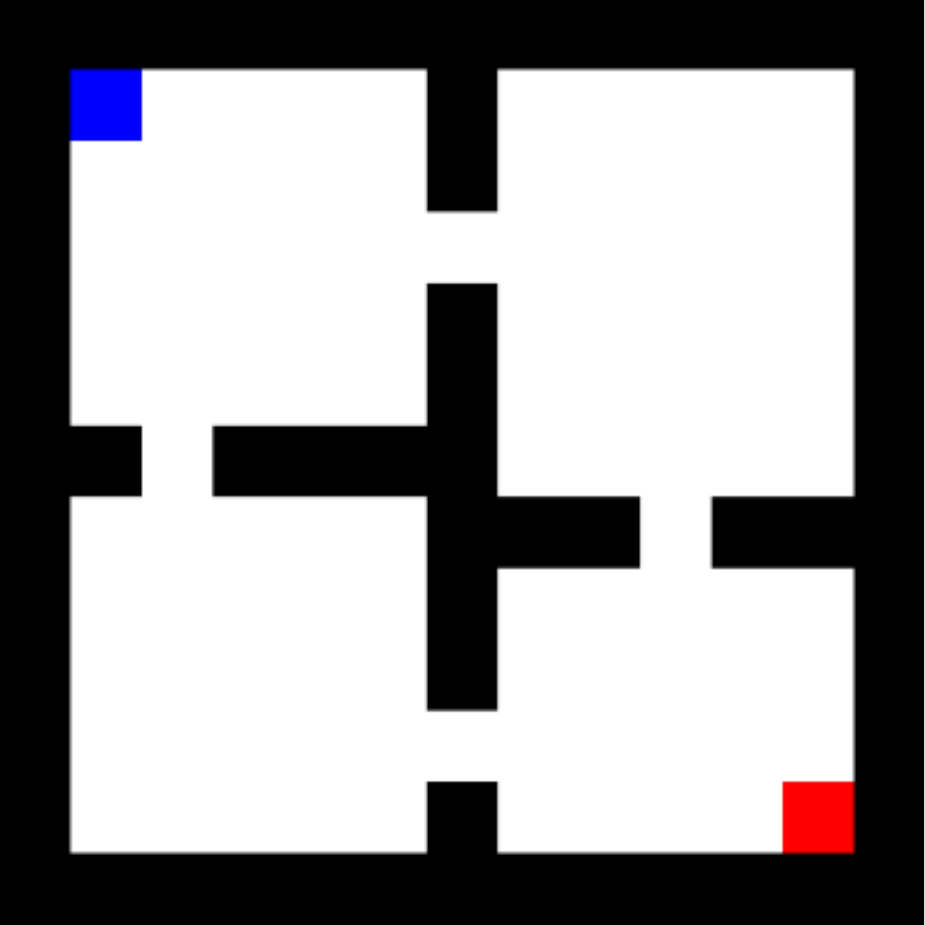}
		\caption{Four-rooms domain.}
		\label{fig:4rooms-vis}
	\end{subfigure}
	\hfill
	\begin{subfigure}[b]{0.49\linewidth}
			\centering
			\includegraphics[width=\linewidth]{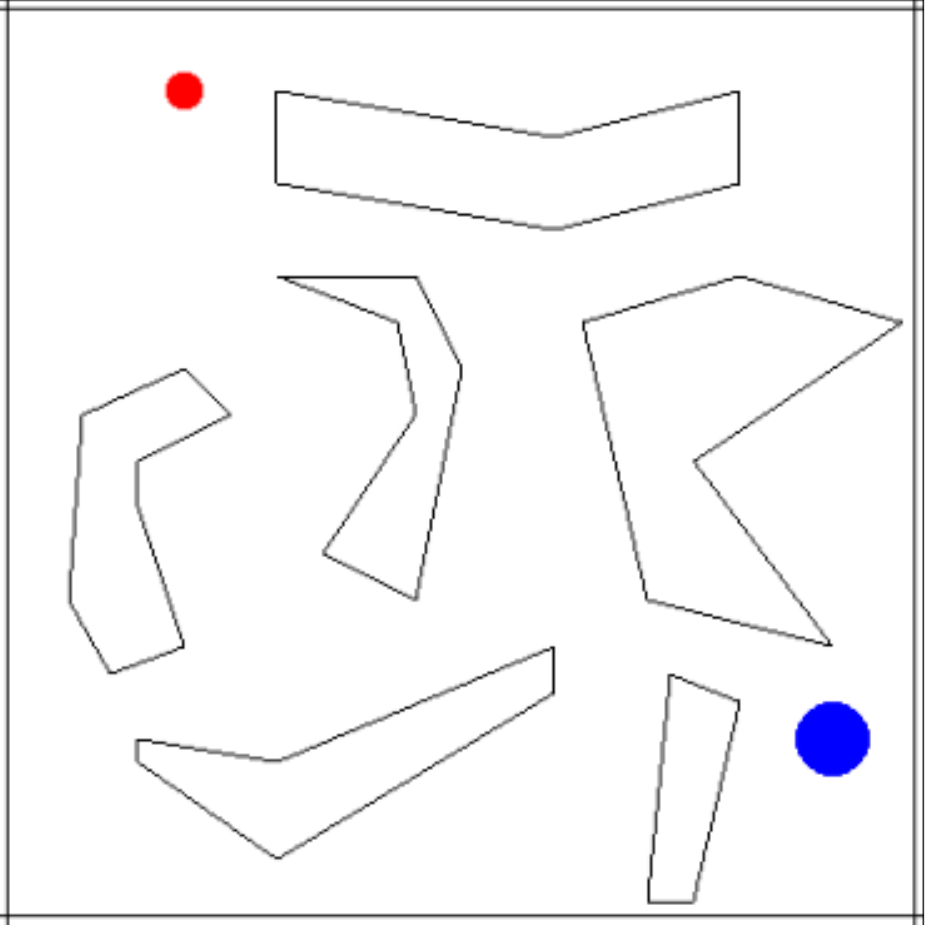}
			\caption{Pinball domain.}
			\label{fig:pinball-vis}
	\end{subfigure}
	\begin{subfigure}[b]{\linewidth}
		\centering
		\includegraphics[width=0.6\linewidth]{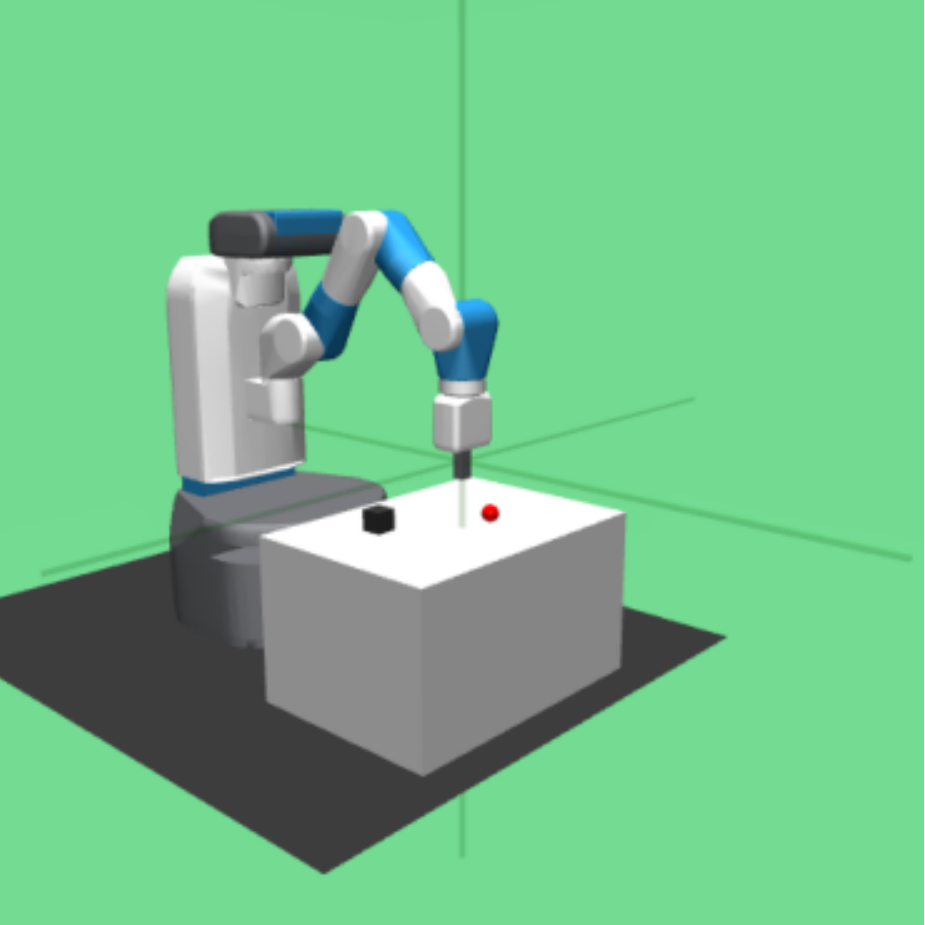}
		\caption{Pick and place domain.}
		\label{fig:pickplace-vis}
\end{subfigure}
    \caption{Three domains.}
    \label{fig:domains}
\end{figure}

As mentioned above, we conducted experiments to evaluate our proposed method in three domains: four-rooms, pinball, and pick and place.
The four-rooms and pinball domains are navigation tasks in which an agent travels to its goal with the shortest number of steps. 
The pick and place domain is a robotics task. A robot aims to pick up an object and bring it to a target space.
Fig.~\ref{fig:domains} shows images of the three domains.

We compared human subgoal-based reward shaping (HSRS) with three other methods. They were a baseline RL algorithm, random subgoal potential-based reward shaping (RSRS), and a naive subgoal reward method (NRS). 
A separate baseline RL algorithm was implemented for each domain. This is to show that our method can be adapted to various RL algorithms.
The different baseline RL algorithms are used each domain to show the availability for several RL algorithms. 
The reward shaping method, HSRS, RSRS, and NRS were implemented in the baseline algorithm. HSRS used the participants' subgoals described in Section~\ref{sec:user_study}. 
% RSRS randomly selected two states as subgoals from the whole state space and used them. Fig.~\ref{fig:random-subgoal-dist} shows the random subgoal distributions of the four-rooms and pinball domains, respectively.

% \begin{figure}[tb]
%     \begin{subfigure}[b]{\linewidth}
%         \centering
%         \includegraphics[width=0.5\linewidth]{misc/4rooms-2sg-random.pdf}
%         \caption{Four-rooms domain}
%         \label{fig:4rooms-random}
%     \end{subfigure}
%     \hfill
%     \begin{subfigure}[b]{\linewidth}
%         \centering
%         \includegraphics[width=0.5\linewidth]{misc/pinball-2sg-random.pdf}
%         \caption{Pinball domain}
%         \label{fig:pinball-random}
%     \end{subfigure}
%     \caption{Random subgoal distribution}
% 	\label{fig:random-subgoal-dist}
% \end{figure}

Each domain had 20 subgoals the same as the number of participants' subgoals. For pick and place, we generated two random observations with ranges as subgoals. The subgoals of pinball and pick and place had these ranges because both domains had a continuous state space.
In NRS, the function $\Phi(s)$ in potential-based reward shaping generates a scalar value $\eta$ just when an agent visits a subgoal state. 
The potential function is written formally as follows.
\begin{eqnarray}
	\Phi(s) = \left \{\begin{array}{ll}
		\eta & s = sg \\
		0 & s \neq sg
	\end{array}
	\right.
\end{eqnarray}
The difference from our method is that the positive potential is only for subgoals.
We evaluated the learning efficiency with the time to threshold and the asymptotic performance~\cite{Taylor:2009:TLR:1577069.1755839} in terms of human subgoal knowledge effects. The time to threshold was the number of episodes to get below a predefined number of threshold steps. The asymptotic performance was the final performance of learning. All the experiments were conducted with a PC [Ryzen 9 5950X, 16 cores(3.4GHz), 128 GB of memory].

\subsection{Navigation in Four-rooms Domain}
The four-rooms domain had four rooms, and the rooms were connected by four hallways. The four-rooms domain is a common RL task. In this experiment, learning consisted of 1000 episodes. An episode was a trial run until an agent reached a goal state successfully or when 1000 state actions ended in failure. A state was expressed as a scalar value labeled through all states. An agent could select an action from up, down, left, and right. The agent's action failed with a probability of 0.33, and another random action was then performed. A reward of +1 was generated when the agent reached a goal state. The start state and goal state were placed at a fixed position. The agent repeated learning 100 times.\par

\subsubsection{Algorithm Setup}
SARSA is a basic RL algorithm. We used SARSA as a baseline RL algorithm. The value function was in table form. The policy was soft-max. We set the learning rate to 0.01 and the discount rate to 0.99. The function $c(h)$ counted the number of subgoals in a sequence in order from the front of a state sequence $h$. We set $\eta$ to 0.01 for HSRS, RSRS and NRS after grid search was performed on grids of 0.01, 0.1, 1, 10, and 100.

\subsubsection{Subgoals}
\begin{figure}[tb]
    \begin{subfigure}[b]{0.49\linewidth}
        \centering
        \includegraphics[width=\linewidth]{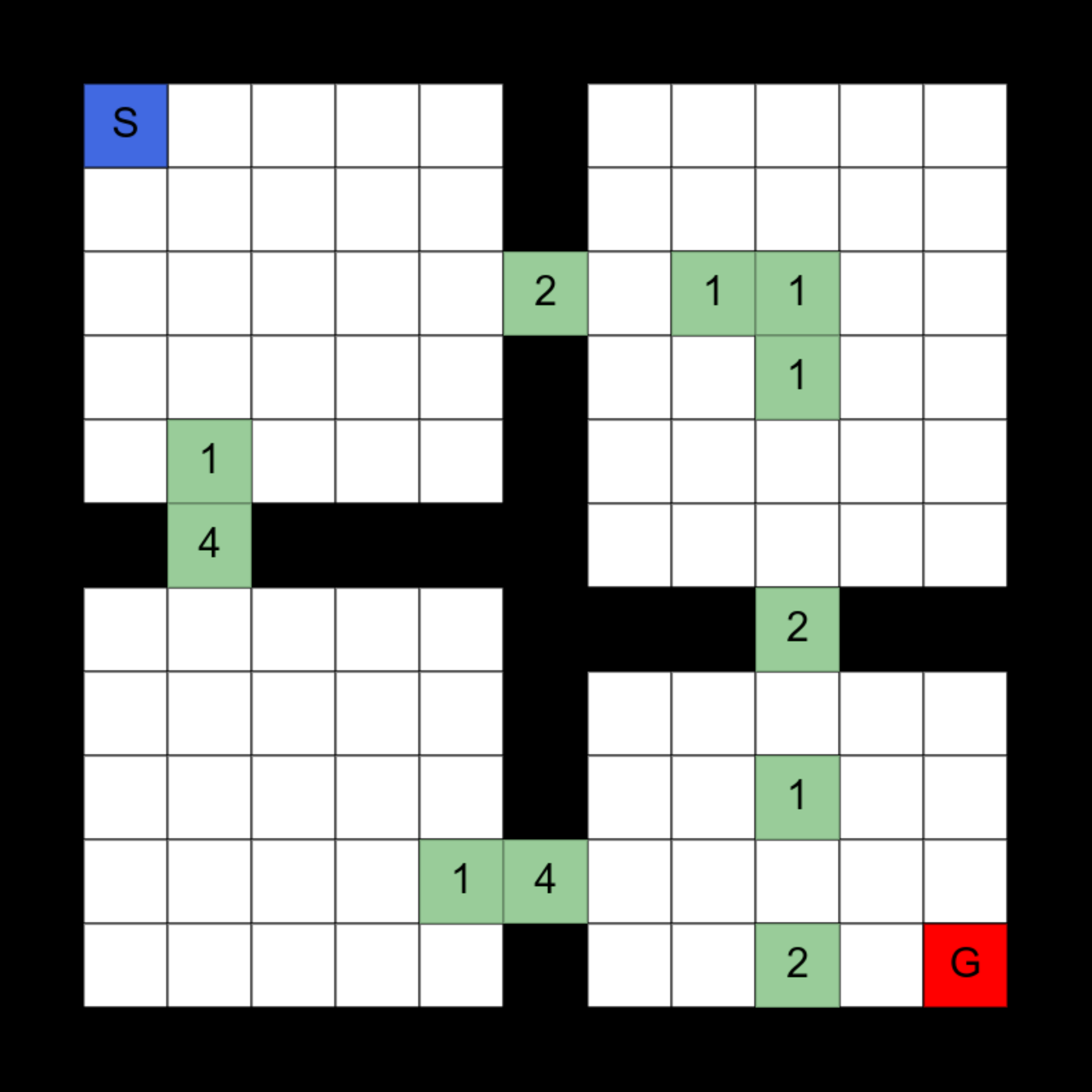}
        \caption{Participants' one.}
        \label{fig:participants-4rooms}
    \end{subfigure}
    \hfill
    \begin{subfigure}[b]{0.49\linewidth}
        \centering
        \includegraphics[width=\linewidth]{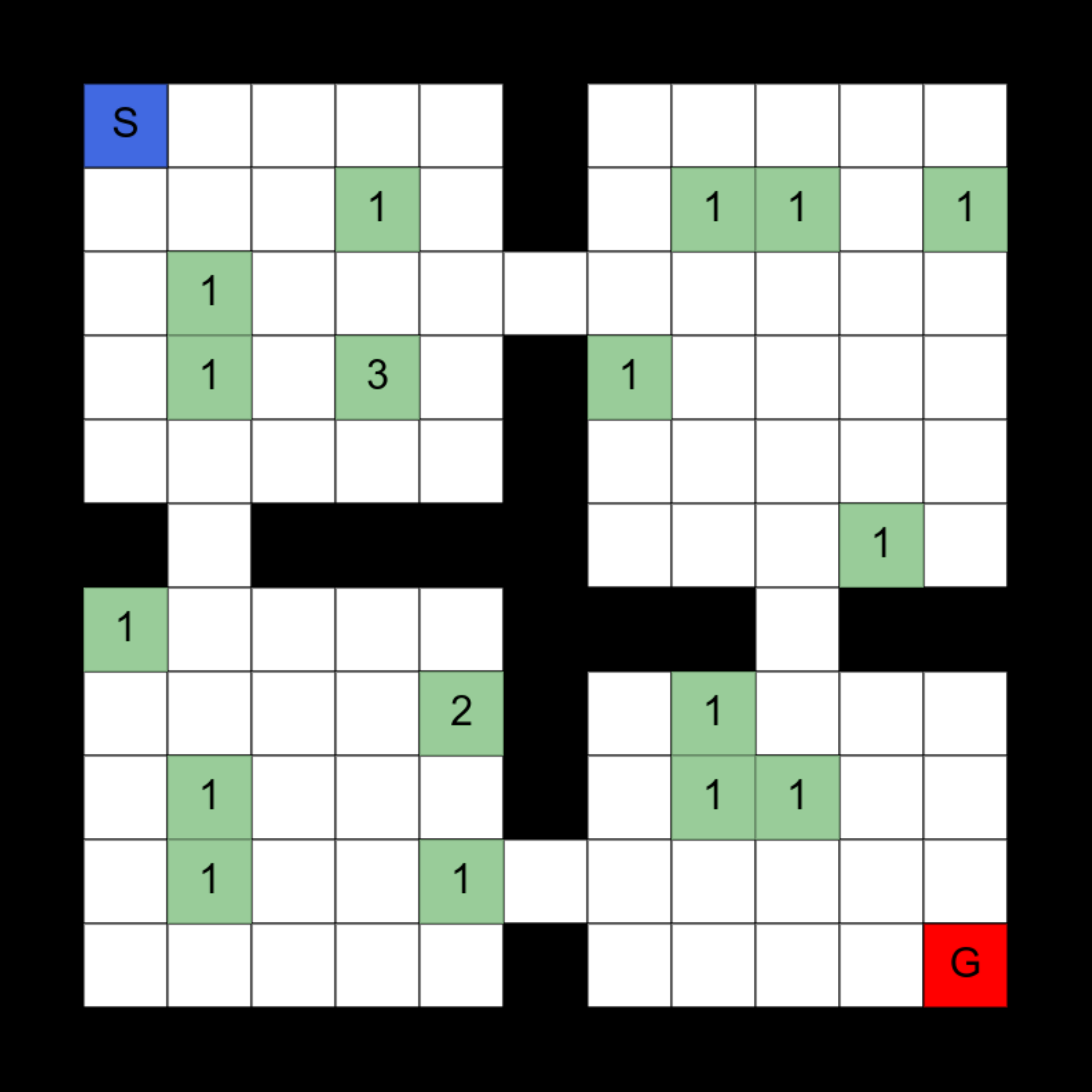}
        \caption{Random one.}
        \label{fig:random-4rooms}
    \end{subfigure}
    \caption{Subgoal distributions of a four-rooms domain.}
	\label{fig:subgoal-dist-4rooms}
\end{figure}

Fig.~\ref{fig:subgoal-dist-4rooms} shows the subgoal distribution acquired from the participants and the random subgoal distribution in the four-rooms domain.

The numbers in the cells stand for the frequencies of the subgoals in Fig.~\ref{fig:subgoal-dist-4rooms}; higher values are in yellow and lower values are in orange. The number in a cell was the frequency at which the participants selected that cell as a subgoal.  These subgoals are totally ordered subgoals. As shown in Fig.~\ref{fig:participants-4rooms}, participants tended to set the subgoals in the hallways. 
In Fig.~\ref{fig:participants-4rooms}, the color of the start point, the goal, and subgoals are red, blue, and green, respectively.
HSRS and NRS used the subgoals as shown in Fig.~\ref{fig:participants-4rooms}, and RSRS used as shown in Fig.~\ref{fig:random-4rooms}.

\subsubsection{Results}
\begin{figure}[tb]
	\includegraphics[width=\linewidth]{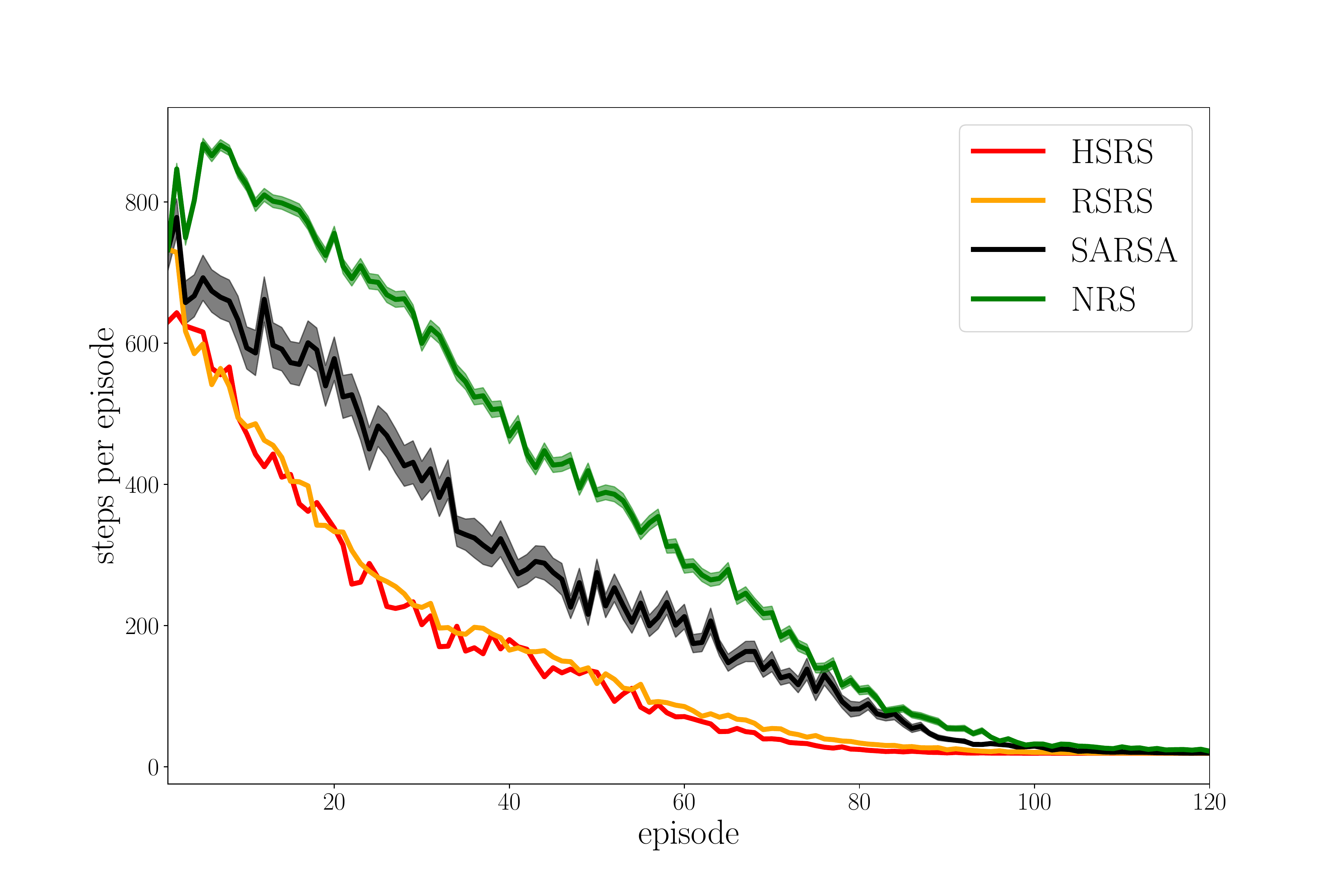}
	\caption{Learning curves compared with three methods.}
	\label{fig:learning-curves}
\end{figure}

We show the results of the learning experiment. Fig.~\ref{fig:learning-curves} shows the learning curves of our proposed method and the three other methods. 

We plotted HSRS with an average totaling 1000 learnings over all participants. RSRS was also averaged by 1000 learnings over all participants. SARSA was averaged by 100 learnings because it does not have the patterns of subgoals. NRS had almost the same conditions as HSRS. HSRS seemed to have the fewest steps for almost all episodes. The performance of RSRS was close to HSRS, but worse than HSRS. NRS had the worst performance. This shows the difficulty of transformation from subgoals into an additional reward function. We also performed an ANOVA for the time to reach the threshold among HSRS, RSRS, SARSA, and NRS. The Holm-Bonferroni method was used as a sub effect test. We set the thresholds to 500, 300, 100, and 50 steps. Table \ref{tab:mean-sd} shows the mean episodes and the standard deviations of the compared methods for each threshold step.

\begin{table}[tb]
	\caption{Mean and standard deviation:Mean(S.D.).}
	\label{tab:mean-sd}
	\begin{center}
	\begin{tabular}{rccccccc}
		\toprule
		Thres.&HSRS&RSRS&SARSA&NRS\\
		\midrule
		500 & 2.17(1.54) &  3.12(2.43) & 3.2(2.61) & 6.45(45.1) \\
		300 & 4.27(3.07) & 5.48(3.73) & 5.69(3.67) &35.3(156) \\
		100 & 18.8(9.42) & 18.7(10.1) & 27.7(11.9) & 175(344) \\
		50 & 35.9(11.7) & 37.7(12.4) & 53.9(17.8) & 376(444)\\
		\bottomrule
	\end{tabular}
	\end{center}
\end{table}

Table \ref{tab:time-to-threshold} shows the results of the ANOVA and the sub effect tests for HSRS, RSRS, SARSA and NRS.

\begin{table}[tb]
	\caption{Summary of ANOVA and sub-effect tests for time(episode) to threshold steps. }
	\label{tab:time-to-threshold}
	\begin{center}
	\begin{tabular}{rcccc}
		\toprule
		Thres. & HSRS\verb|<|NRS&RSRS\verb|<|NRS&SARSA\verb|<|NRS&otherwise\\
		\midrule
		500 & * & * & * &n.s.\\
		300 & * & * & * &n.s. \\
		100 & * & * & * &n.s. \\
		50 & * & * & * &n.s. \\
		\bottomrule
	\end{tabular}
	\end{center}
\end{table}

As shown in Table\ref{tab:mean-sd} and Table\ref{tab:time-to-threshold}, HSRS, RSRS, and SARSA were significantly lower than NRS but did not have significant differences between the three of them. From Table~\ref{tab:mean-sd}, HSRS shortened the number of learnings by up to 18 episodes from SARSA.
Table~\ref{tab:asym-pinball} shows the asymptotic performances of all four methods.

\begin{table}[tb]
    \centering
    \caption{Asymptotic performances in four-rooms domain. Ind. is indicator, and S.D. is standard deviation.}
    \label{tab:asym-4rooms}
    \begin{tabular}{rcccc}
        \toprule
        Ind.&HSRS & RSRS & SARSA & NRS \\
        \midrule
        Mean&19.1 & 19.1 & 24.9 & 552 \\
        S.D. & 0.40 & 0.41 & 13.5 & 455 \\
        \bottomrule
    \end{tabular}
\end{table}

There were no significant differences among the four methods in terms of asymptotic performance. 

\subsection{Navigation in Pinball Domain}
The navigation task in the pinball domain involves moving a ball to a designated target by giving it velocity. The pinball domain makes it difficult for humans to control the ball because it needs them to control the ball with delicate actions. Delicate control is often necessary in the control domain. It is easier to give a subgoal than trajectories in this domain. \par
The difference with the four-rooms domain is the continuous state space over the position and velocity of the ball on the x-y plane. An action space has five discrete actions, four types of force and no force.  In this domain, a drag coefficient of 0.995 effectively stops ball movements after a finite number of steps when the no-force action is chosen repeatedly; collisions with obstacles are elastic. The four types of force are 
up, down, right, left on a plane. An episode terminates with a reward of +10,000 when the agent reaches the goal. Interruption of any episode occurs when the episode takes more than 10000 steps. Learnings of the methods are repeated 100 times from scratch, and we used the results to evaluate the learning efficiency.

\subsubsection{Algorithm Setup}
We used the actor-critic algorithm(AC) \cite{NIPS1999_1786} as the baseline RL algorithm.
AC is a basic RL algorithm that directly optimizes a policy with parameters using the value function. A policy with parameters is called the actor part, and a value function is the critic part. Both parts needed to update parameters. We used AC as the critic with linear function approximation over a Fourier basis~\cite{Konidaris11a} of order 3. 
A subgoal had only a center position and a radius in this domain. The radius was the same as that of the target. We assumed that a subgoal was achieved when the ball entered a circle with the center position and radius at any velocity. We used a soft-max policy as the actor. The learning rates were set to 0.01 for both the actor and the critic. The discount factor was set to 0.99. 
The parameters for HSRS, RSRS, NRS, and $\eta$ were set to 100 after a grid search on grids of 10, 100, 1,000, and 10,000. The function $c(h)$ was almost the same as in the four-rooms domain excluding the judgement of whether a state is a subgoal. We compared HSRS with the three other methods in terms of learning efficiency with the time to threshold. We defined this threshold in this domain as the number of episodes required to reach the steps. 

\subsubsection{Subgoals}
\begin{figure}[tb]
    \begin{subfigure}[b]{0.49\linewidth}
        \centering
        \includegraphics[width=\linewidth]{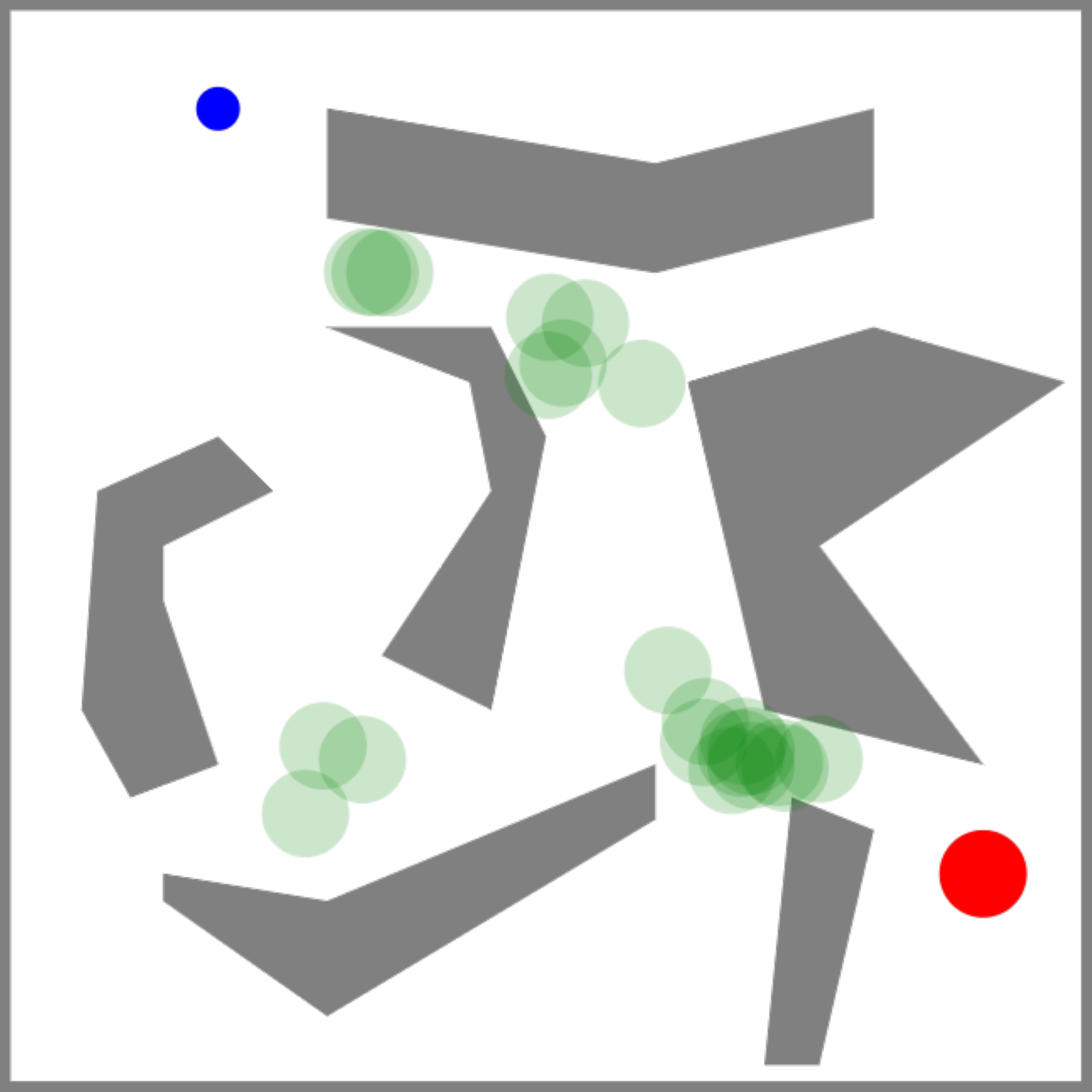}
        \caption{Participants' one.}
        \label{fig:participants-pinball}
    \end{subfigure}
    \hfill
    \begin{subfigure}[b]{0.49\linewidth}
        \centering
        \includegraphics[width=\linewidth]{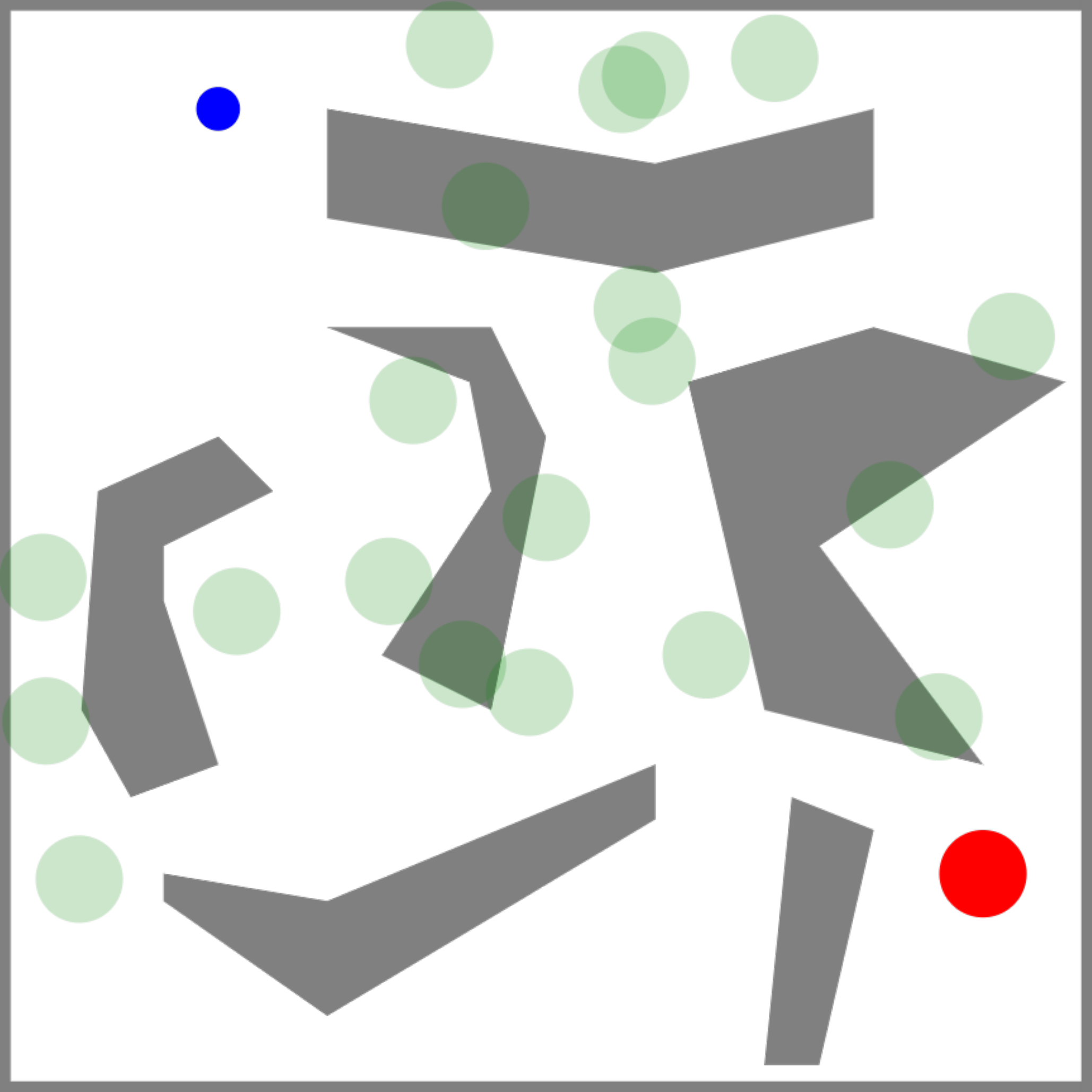}
        \caption{Random one.}
        \label{fig:random-pinball}
    \end{subfigure}
    \caption{Subgoal distributions of a pinball domain.}
	\label{fig:subgoal-dist-pinball}
\end{figure}

% Participants' subgoalの特徴についてここで説明。ランダムの生成の仕方を説明。
Fig.~\ref{fig:subgoal-dist-pinball} shows the subgoal distribution acquired from the participants and the random subgoal distribution in the pinball domain.

The participants focused on four regions of branch points to set subgoals in comparison with random subgoals from Fig.~\ref{fig:participants-pinball} and Fig.~\ref{fig:random-pinball}.
HSRS and NRS used the subgoals as shown in Fig.~\ref{fig:participants-pinball}, and RSRS used those as shown in Fig.~\ref{fig:random-pinball}. 

\subsubsection{Results}
\begin{figure}[tb]
	\begin{center}
	\includegraphics[width=\linewidth]{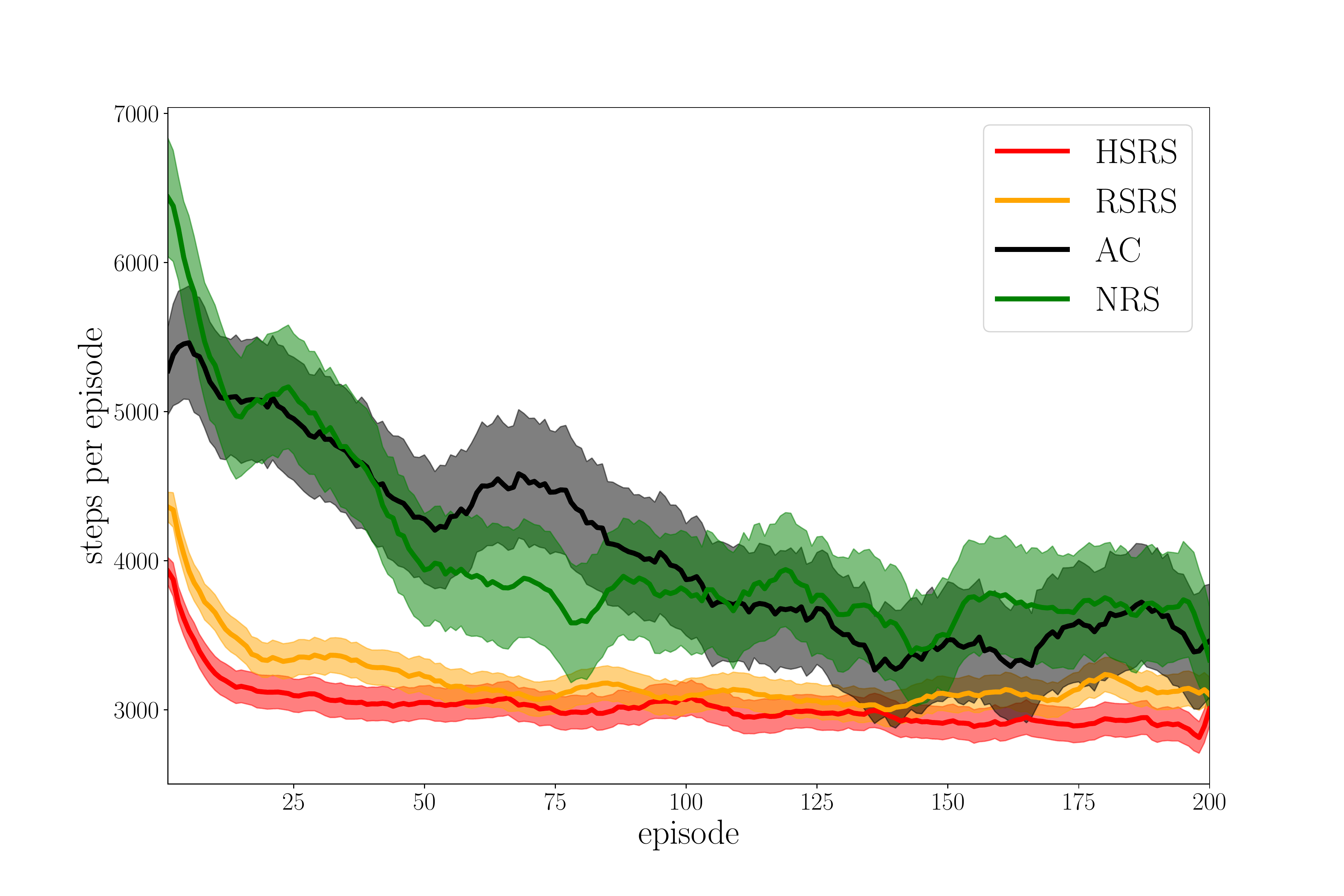}
	\caption{Learning curve in pinball domain.}
	\label{fig:lc-pinball}
	\end{center}
\end{figure}

Fig.~\ref{fig:lc-pinball} shows the learning curves. It took an average shift of 10 episodes.

HSRS mostly performed the best of the three methods. The performance of RSRS was close to HSRS, but worse than HSRS. We evaluated the learning efficiency by using the time to threshold. We used each learning result smoothed using the simple moving average method with the number of time periods being 10 episodes, and performed a Student's t-test to confirm the difference among the three methods. 
Table~\ref{tab:meansd-pinball} shows the mean and the standard deviations, and Table~\ref{tab:summary-pinball} summarizes the results for the time to threshold, where Thres. is the number of steps.

\begin{table}[tb]
	\caption{Mean and standard deviation: Mean(SD).}
	\label{tab:meansd-pinball}
	\begin{center}
		\begin{tabular}{rcccc}
			\toprule
			Thres.&HSRS&RSRS&AC&NRS\\
			\midrule
			3000& 15.0 (32.7) & 23.8 (45.7)& 51.7 (67.1) & 39.8 (50.2)  \\
			2000& 28.0 (38.8) & 33.2 (46.8)& 65.4 (67.6) & 59.0 (58.6) \\
			1000& 63.7 (60.0) & 72.4 (72.8) & 101 (71.3) &92.0 (69.2) \\
			500& 140 (67.0) &141 (14.5)& 163 (54.8) &158 (61.0) \\
			\bottomrule
		\end{tabular}
	\end{center}
\end{table}

\begin{table}[tb]
	\caption{Summary of ANOVA and sub effect tests for time(episode) to threshold steps. Otherwise is abbreviated to o.w., and n.s. means not significant.}
	\label{tab:summary-pinball}
	\begin{center}
	\begin{tabular}{rcccc}
		\toprule
		Thres.&\{HSRS, RSRS\}\verb|<|AC&HSRS\verb|<|NRS&RSRS\verb|<|NRS& o.w.\\
		\midrule
		3000 & * & * & * &n.s.\\
		2000 & * & * & * &n.s. \\
		1000 & * & * & n.s. &n.s. \\
		500 & n.s. & n.s. & n.s. &n.s. \\
		\bottomrule
	\end{tabular}
	\end{center}
\end{table}

From Table~\ref{tab:summary-pinball}, HSRS and RSRS reached steps of 3000, 2000, and 1000 less than AC, which was significant. There was no significant difference between HSRS and RSRS for any of the thresholds. Table~\ref{tab:asym-pinball} shows the asymptotic performances of all four methods.

\begin{table}[tb]
    \centering
    \caption{Asymptotic performances in pinball domain. Ind. is indicator, and S.D. is standard deviation.}
    \label{tab:asym-pinball}
    \begin{tabular}{rcccc}
        \toprule
        Ind.&HSRS & RSRS & AC & NRS \\
        \midrule
        Mean&3008 & 3629 & 3625 & 3672 \\
        S.D. & 2678 & 2925 & 3014 & 3042 \\
        \bottomrule
    \end{tabular}
\end{table}

There were no significant differences among the four methods in terms of asymptotic performance. 
Our method with random and human subgoals lead to more efficient learning than the basic RL algorithm in the pinball domain. 

\subsection{Pick and Place}
We used a fetch environment based on the 7-DoF Fetch robotics arm of OpenAI Gym\cite{1606.01540}. In pick and place, the robotic arm learns to grasp a box and move it to a target position\cite{DBLP:journals/corr/abs-1802-09464}. We converted the original task into a single-goal reinforcement learning framework because potential-based reward shaping is not developed to deal with the multiple goals~\cite{Ng+HR:1999}. The dimension of observation is larger than the previous navigation tasks, and the action is continuous. An observation is 25-dimensional, and it includes the Cartesian position of the gripper and the object as well as the object's position relative to the gripper. The reward function generates a reward of -1 every step and a reward of 0 when the task is successful. In \cite{DBLP:journals/corr/abs-1802-09464}, the task is written about in detail.

\subsubsection{Algorithm Setup}
We compared HSRS with RSRS, NRS, and DDPG~\cite{journals/corr/LillicrapHPHETS15} in terms of learning efficiency with the time to threshold and asymptotic performance. HSRS, RSRS, and NRS used DDPG as a base. We defined this threshold in the task as the number of epochs required to reach the designated success rate. The asymptotic performance was the average success rate between 190 and 200 epochs. Ten workers stored episodes and calculated the gradient simultaneously in an epoch. HSRS and NRS used ordered subgoals provided by five participants. RSRS used subgoals randomly generated. The learning in 200 epochs took several hours. We used an OpenAI Baselines\cite{baselines} implementation for DDPG with default hyper-parameter settings. We built the hidden and output layers of the value network over abstract states with the same structure as the Q-value network. We excluded the action from the input layer. The input of the network was only the observation of subgoal achievement, and the network learned from the discount accumulated reward until subgoal achievement. A subgoal was defined from the information in the observation. We set the margin to $\pm{0.01}$ to loosen the severe condition to achieve subgoals. $\eta$ was set as 1 for HSRS, RSRS, and NRS. The learning repeated 10 times, and the results were averaged over 10 learnings.

\subsubsection{Subgoals}
\begin{figure}[tb]
	\begin{subfigure}[b]{0.49\linewidth}
			\centering
			\includegraphics[width=\linewidth]{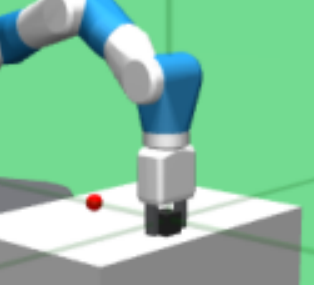}
			\caption{First subgoal.}
			\label{fig:first-subgoal}
	\end{subfigure}
	\hfill
	\begin{subfigure}[b]{0.49\linewidth}
			\centering
			\includegraphics[width=\linewidth]{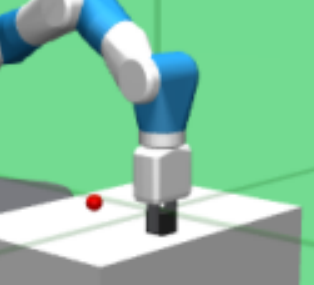}
			\caption{Second subgoal.}
			\label{fig:second-subgoal}
	\end{subfigure}
	\caption{Example of subgoals from participant in a pick and place domain.}
	\label{fig:subgoals-pick-place}
\end{figure}

All five participants determined the subgoal series, the first subgoal was reaching the location available to grasp the object, and the second subgoal was grasping the object. Fig.~\ref{fig:subgoals-pick-place} shows the example of the subgoal series in pick and place domain. 

\subsubsection{Results}
\begin{figure}[tb]
	\centering
	\includegraphics[width=\linewidth]{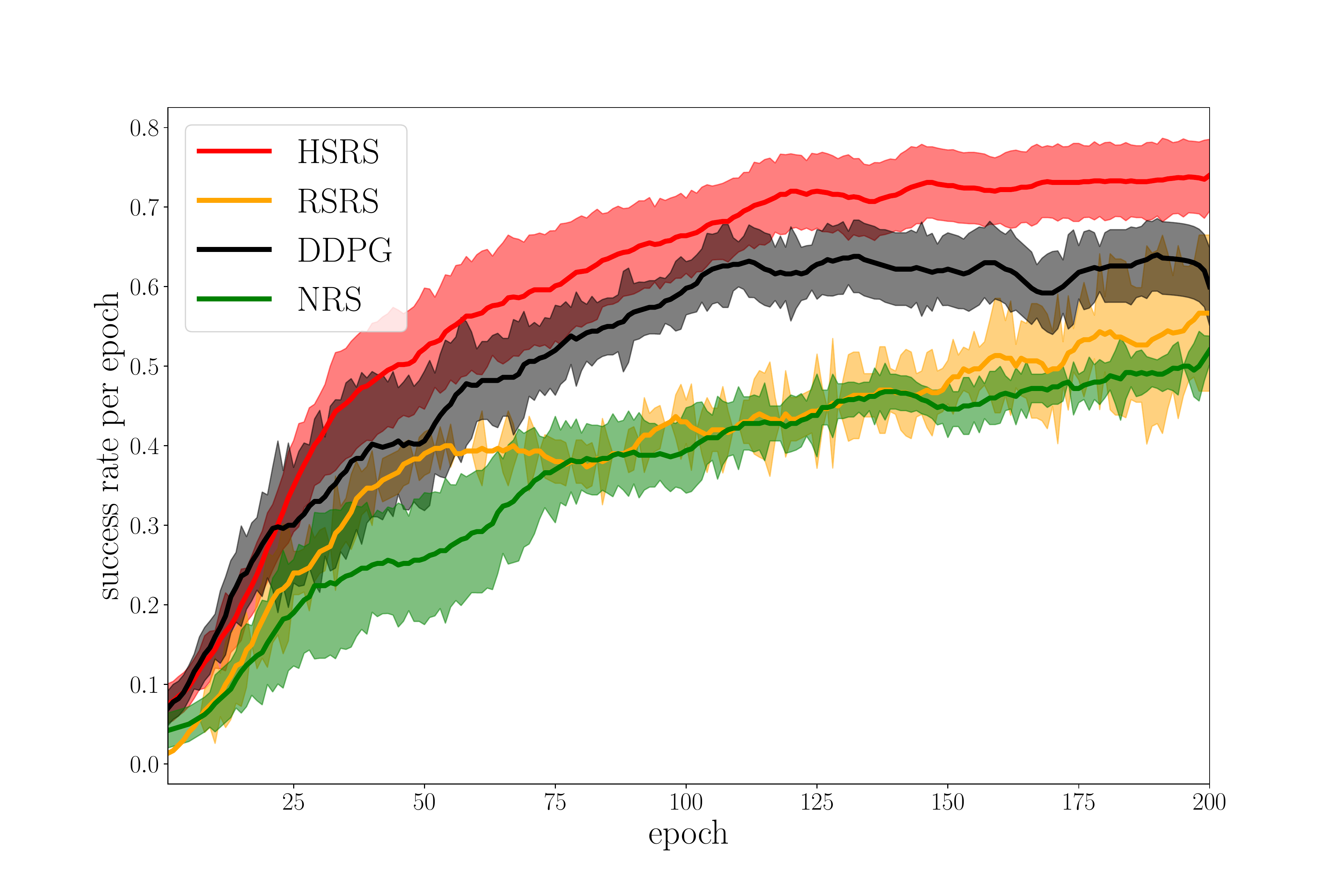}
	\caption{Learning curves in pick and place task.}
	\label{fig:lc-robotics}
\end{figure}

We used the subgoal series for the input of our method. Fig.~\ref{fig:lc-robotics} shows the learning curves of HSRS, RSRS, NRS, and DDPG. 

As shown as Fig.~\ref{fig:lc-robotics}, the results were averaged across five learnings, and the shaded areas represent one standard error. The random seeds and the locations of the goal and object were varied every learning. HSRS worked more effectively than DDPG, especially after about the 75th epoch. NRS had the worst performance through almost all epochs.

Table~\ref{tab:mean-sd-pickplace} shows the mean episodes and the standard deviations of the compared methods for each threshold steps.

\begin{table}[tb]
	\caption{Mean and standard deviation in pick and place:Mean(S.D.).}
	\label{tab:mean-sd-pickplace}
	\begin{center}
	\begin{tabular}{rccccccc}
		\toprule
		Thres.&HSRS&RSRS&DDPG&NRS\\
		\midrule
		0.2 & 26.3 (22.3) &  27.1 (6.77) & 28.2 (26.4) & 34.3 (18.7) \\
		0.4 & 42.7 (27.2) & 87.0 (49.6) & 56.6 (37.5) &75.9 (43.5) \\
		0.6 & 75.1 (56.7) & 175 (46.4) & 110 (63.7) & 189 (26.2) \\
		0.8 & 156 (57.6) & 185 (35.7) & 187 (22.9) & 200 (0)\\
		\bottomrule
	\end{tabular}
	\end{center}
\end{table}

Table~\ref{tab:asym-pickplace} shows the asymptotic performances of all four methods.

\begin{table}[tb]
    \centering
    \caption{Asymptotic performances in pick and place. Ind. is indicator, and S.D. is standard deviation.}
    \label{tab:asym-pickplace}
    \begin{tabular}{rcccc}
        \toprule
        Ind.& HSRS & RSRS & DDPG & NRS \\
        \midrule
        Mean & 0.73 & 0.53 & 0.63 & 0.51 \\
        S.D. & 0.16 & 0.17 & 0.13 & 0.06 \\
        \bottomrule
    \end{tabular}
\end{table}

HSRS was significantly different from RSRS and NRS at 0.6 in terms of the time to threshold. For the rest, there were no differences in both the time to threshold and the asymptotic performance, but HSRS was the best compared with the other three methods. 

In this section, we evaluated our method for four-rooms, pinball, and pick and place. In all domains, HSRS performed better than the other three methods in terms of the time to threshold and the asymptotic performance. There was, in particular, a significant difference between HSRS and AC as the baseline RL algorithm for pinball. The performance of RSRS was close to HSRS in the four-rooms and pinball domains. In contrast, RSRS performed the second worst of all for pick and place. The results showed the subgoal-based reward shaping with human knowledge of subgoal improved the learning efficiency from the baseline RL algorithm, and the human knowledge was more useful for SRS than random. The naive reward shaping made the learning efficiency worse than the baseline RL algorithm, so it is not easy for designers to incorporates the knowledge of subgoal over PBRS. This is why we proposed SRS, and SRS could improve the learning efficiency just by providing the subgoal series.

\section{Discussion}
\label{sec:discussion}
% TODO 最も性能が良くなったサブゴールがどれかをここで分析しても良いかも
% TODO 提案法に特化した特別なGUIが必要で今後開発していくという主旨で，あと1 paragraph を加筆する．First-pseson's view/birds' eyes view にも少し触れる．文献が2，3つ欲しいです．
\subsection{Analyzing Performances}
% Modify contents to match the results.
There was a small difference between HSRS and RSRS in the four-rooms domain from Fig.~\ref{fig:learning-curves} and from Fig.~\ref{fig:lc-pinball}. The difference between HSRS and RSRS was large in the pick and place domain. Approximately 65\% of states generated randomly were in an optimal trajectory for the four-rooms domain. The pinball task seemed to have approximately 20\% of states generated randomly in the optimal trajectory. There is no state generated randomly in an optimal trajectory.
The random subgoals were better in the four-rooms and pinball domain than in the pick and place task. This is because the four-rooms and pinball domain might have more states in an optimal trajectory than the pick and place task. We think that the smaller difference was caused by the four-rooms and pinball domain having most of the states in an optimal trajectory.

Potential-based reward shaping keeps a policy invariant from the transformation of a reward function. From the experimental results in the pinball domain, the asymptotic performances of HSRS were statistically significantly lower than RSRS. 
There was no significant difference in the four-rooms domain. As shown in Fig.~\ref{fig:learning-curves}, the performance was clearly asymptotic at the 121th of 1000 episodes.
The learning was clearly asymptotic as the 200th episode for the pinball domain in Fig.~\ref{fig:lc-pinball}. Since our proposed method is based on PBRS, RSRS converges to the same performance as HSRS if learning continues.
\subsection{Hyper Parameters}
\begin{figure}[tb]
    \begin{subfigure}[b]{\linewidth}
        \centering
        \includegraphics[width=\linewidth]{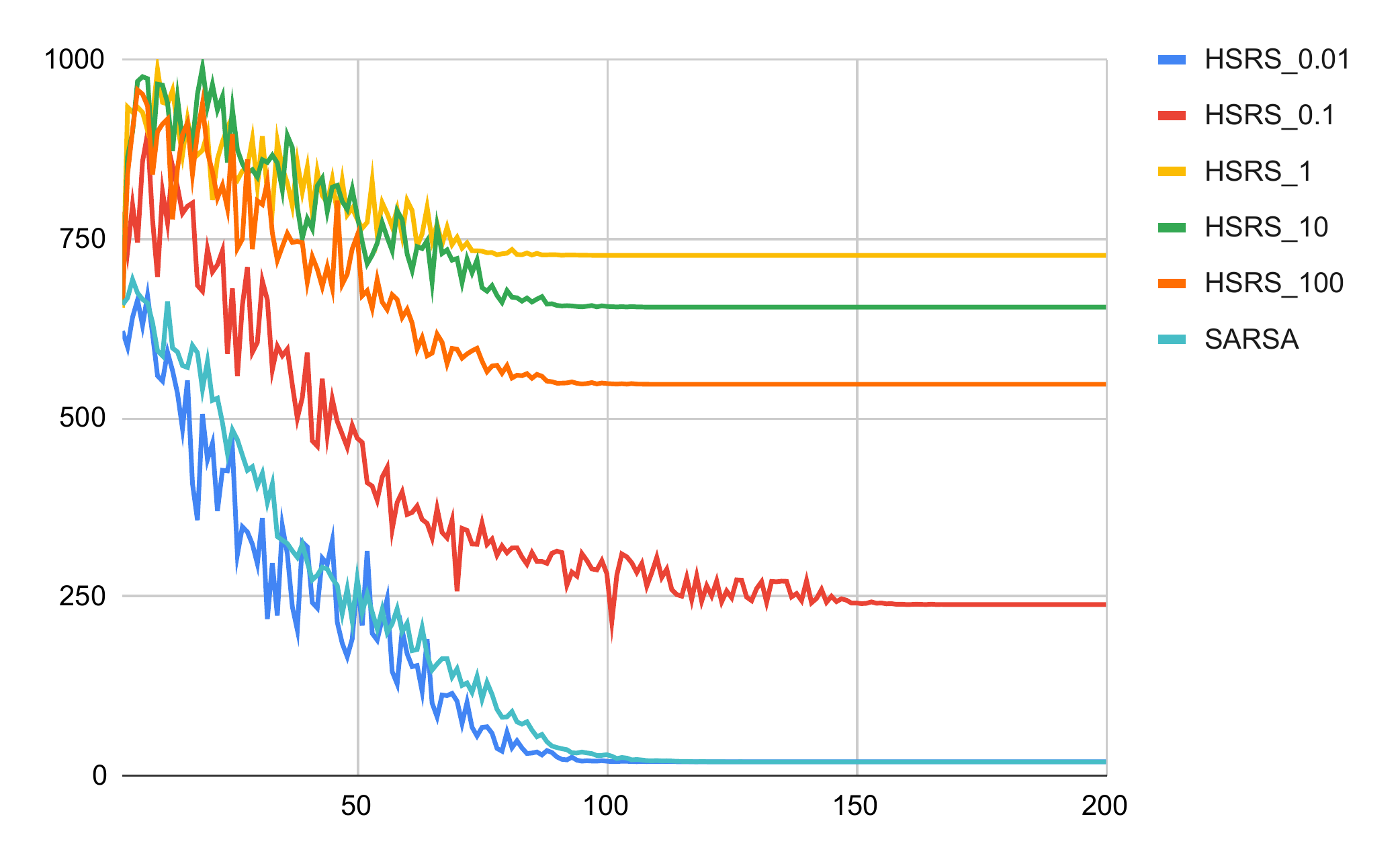}
        \caption{Four-rooms domain.}
        \label{fig:4rooms_hypara}
    \end{subfigure}
    \hfill
    \begin{subfigure}[b]{\linewidth}
        \centering
        \includegraphics[width=\linewidth]{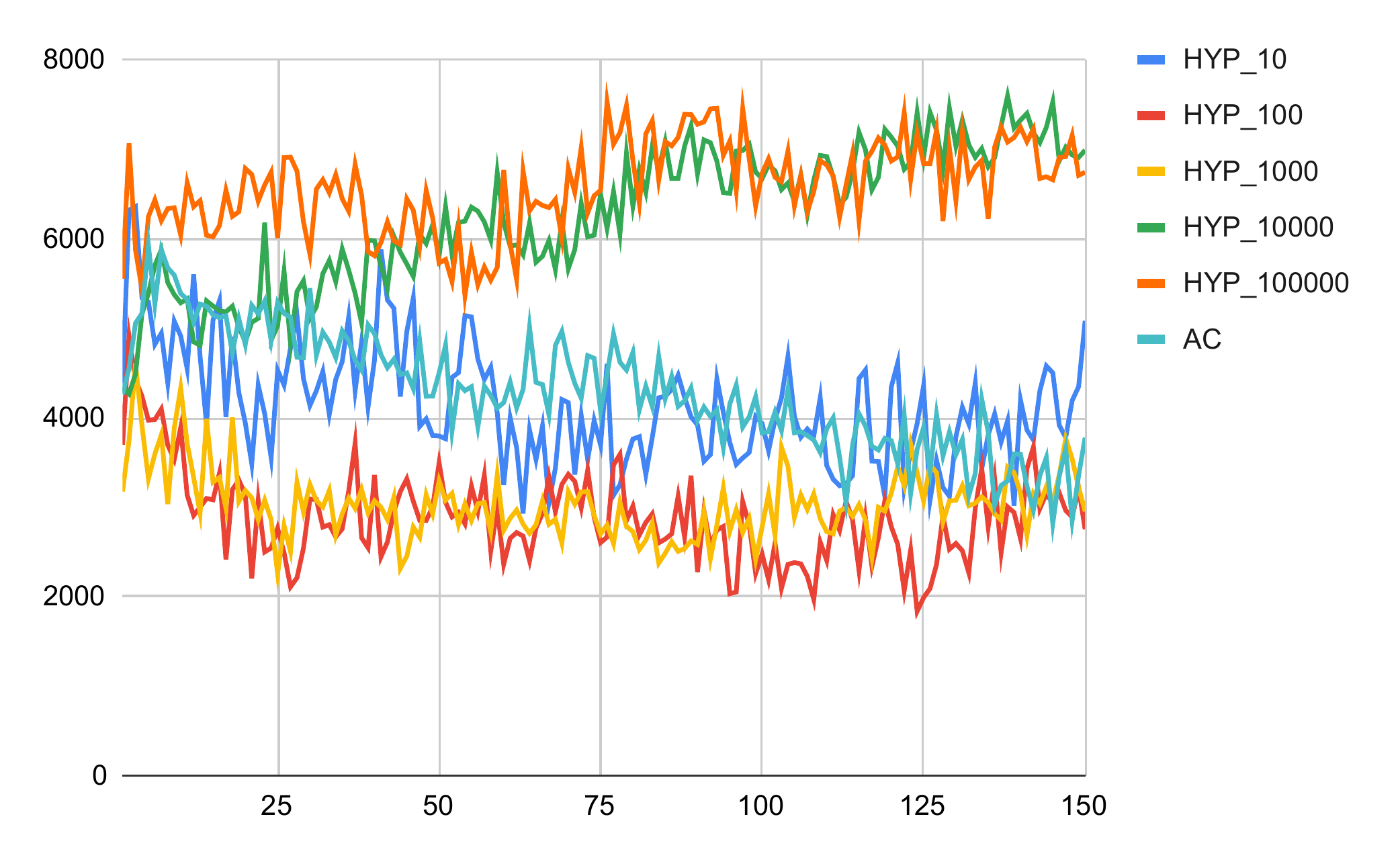}
        \caption{Pinball domain.}
        \label{fig:pinball_hypara}
    \end{subfigure}
    \caption{Results of grid searches.}
	\label{fig:hypara}
\end{figure}

Fig.~\ref{fig:hypara} shows the results of grid searches. In the searches, we used only a single participant's subgoals which were randomly selected, and the number of learnings was a half that of the experimental results. 

Seeing from this figure, our method was significantly sensitive to the hyper parameter $\eta$. A wrong $\eta$ such as $\eta=1$ in the four-rooms domain could deteriorate the performance of the method. This may incur a cost for tuning hyper parameters. To solve this problem, we must fit $\eta$ to optimal value. The environmental rewards of goals were 1 and 10,000 for four-rooms and pinball, respectively. The best $\eta$ was 0.01 for four-rooms and 100 for pinball. Both $\eta$ were one-hundredth of the environmental rewards.  

\subsection{Teaching Knowledge of Subgoals}
% 確立されたサブゴールを与える手法がなくて属人的 -> 実験結果を見て考察 -> Future workにつなげる。
The limitation of subgoals is that no established method to provide them, and it may depend on individual intuitive consideration. For the four-rooms domain, as shown in Fig.~\ref{fig:subgoal-dist-4rooms}, almost all the subgoals were located within the right-top and right-bottom rooms. From this, we think that many participants tended to consider the right-down path as the shortest one. This may mean humans abstractly have a common methodology and preference for giving subgoals. Additionally, there was the interesting observation that half of the participants set a subgoal in the hallways. \par 
For the pinball domain, as shown in Fig.~\ref{fig:subgoal-dist-pinball}, the subgoals were scattered at the gap between objects and at the branch points. They seem to be at the change points of actions on an optimal trajectory. We can consider that the repeated actions are composed in a macro-action. This is a similar approach to~\cite{10.5555/3298483.3298547}. \par
For the pick and place task, five participants provided the same subgoals. The way to provide subgoals was different from the other tasks. We used the descriptive answer-type form. These results almost show the tendency for people to provide similar subgoals exists. \par
We consider that people change a strategy for providing subgoals in response to the task properties, especially environment structure. The test for this hypothesis is future work.  \par
There are other limitations and future work regarding this work. 
We implicitly assumed that a human can set subgoals  more easily than providing the optimal trajectories from a start state to a goal state, which is conventional human knowledge that improves efficiency of reinforcement learning with IRL. We need to verify this assumption by conducting experiments with participants to measure the cognitive load in setting various subgoals and trajectories. We consider that the ease of providing subgoals significantly depends on the kind of subgoals. We plan to conduct a large user study to make clear what is the best task for providing subgoals. 

\section{Conclusion}
\label{sec:conclusion}
In reinforcement learning, learning a policy is time-consuming. We aim for accelerating learning with reward transformation based on human subgoal knowledge. 
The difficulty of integrating subgoal knowledge into the potentials was a problem in improving learning efficiency. We proposed a method by which a human deals with several characteristic states as a subgoal. We defined a subgoal as the goal state in one of the sub-tasks into which a human decomposes a task. 
The main part of our method is an approximation of the optimal value function with subgoals. We collected ordered subgoals from participants and used them for evaluation. We evaluated navigation for four-rooms, pinball, and a pick and place task. The experimental results revealed that our method with human subgoals enabled faster learning compared with the baseline method, and human subgoal series were more helpful than random ones. 
\bibliography{Bibliography.bib}
\bibliographystyle{IEEEtran}

\begin{IEEEbiography}[{\includegraphics[width=1in,height=1.25in,clip,keepaspectratio]{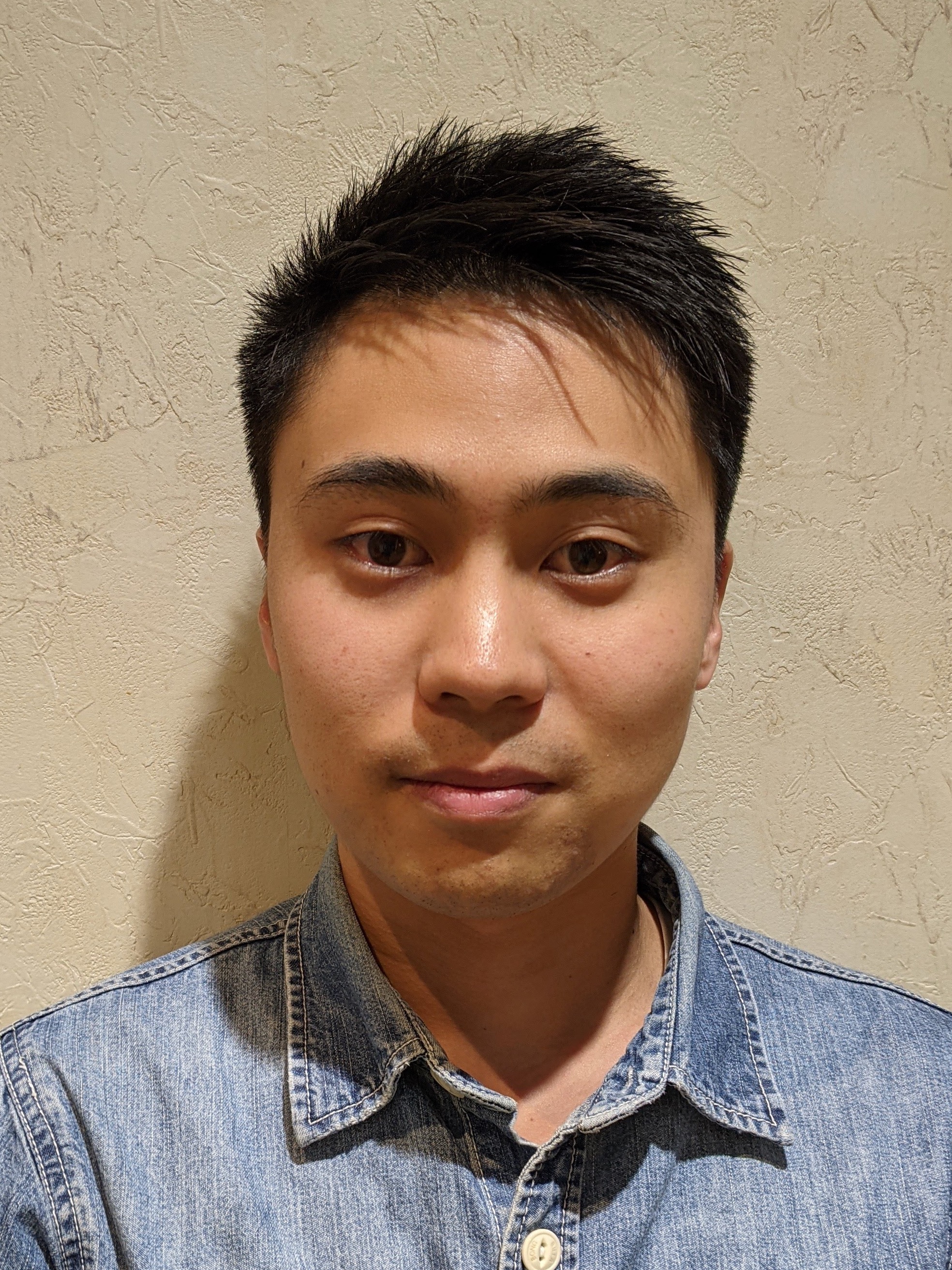}}]{Takato Okudo}
	was born in Osaka, Japan in 1995. He received the B.S. degree in informatics from National Institute of Technology, Nara College in 2018. \par
	Since 2018, he has been a Ph.D. student in the Graduate University for Advanced Studies and a Research Assistant with the National Institute of Informatics. His research interests include machine learning, reinforcement learning, human agent interaction, and human knowledge transfer.
\end{IEEEbiography}

\begin{IEEEbiography}[{\includegraphics[width=1in,height=1.25in,clip,keepaspectratio]{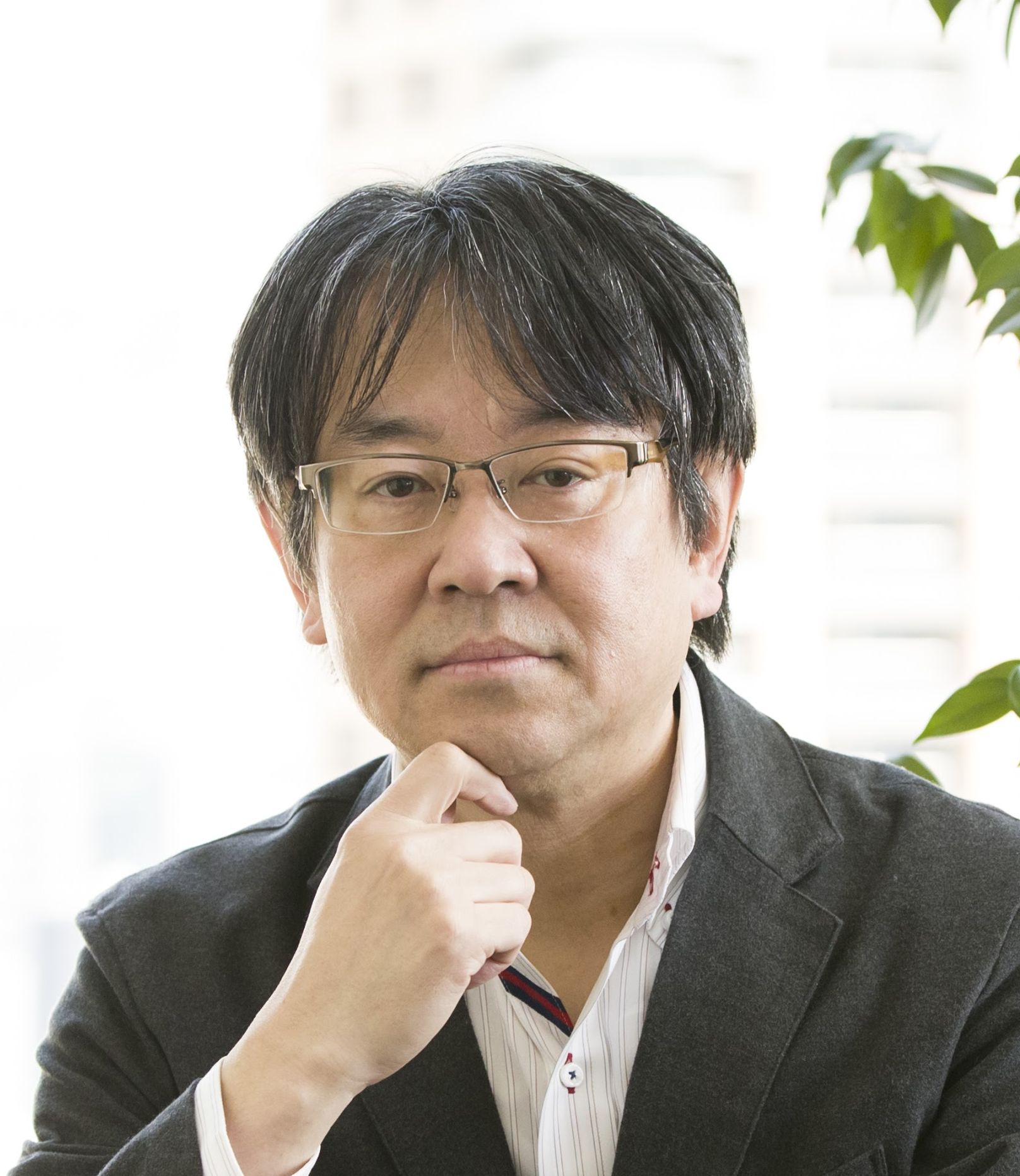}}]{Seiji Yamada} is a professor at the National Institute of Informatics and The Graduate University for Advanced Studies (SOKENDAI). Previously he worked at Tokyo Institute of Technology. He received B.S., M.S. and the Ph.D. degrees in artificial intelligence from Osaka University. His research interests are in the design of intelligent interaction including Human-Agent Interaction, intelligent Web interaction and interactive machine learning.
\end{IEEEbiography}

\EOD

\end{document}